\documentclass[10pt,twocolumn,letterpaper]{article}

\usepackage{cvpr}
\usepackage{times}
\usepackage{epsfig}
\usepackage{multirow}
\usepackage{graphicx}
\usepackage{amsmath}
\usepackage{amssymb}
\usepackage{booktabs}
\usepackage{stmaryrd}
\usepackage{tabularx}
\usepackage{algorithm}
\usepackage{algpseudocode}
\usepackage{pifont}
\usepackage{color, colortbl}
\usepackage[table,xcdraw,dvipsnames]{xcolor}
\usepackage[pagebackref,breaklinks,colorlinks,citecolor=cvprblue]{hyperref}

\definecolor{cvprblue}{rgb}{0.21,0.49,0.74}
\definecolor{Gray}{gray}{1}
\definecolor{my_gray}{gray}{0.95}

\newcommand{\std}[1]{\small{$\pm$ #1}}
\newcolumntype{C}[1]{>{\centering\arraybackslash\hspace{0pt}}p{#1}}

\title{Leveraging Vision-Language Models for Improving \\Domain Generalization in Image Classification}

\author{%
     Sravanti Addepalli \thanks{Equal Contribution. \\ Correspondence to Sravanti Addepalli $<$sravantia@iisc.ac.in$>$, Ashish Ramayee Asokan $<$ashish.ramayee@gmail.com$>$} \quad  Ashish Ramayee Asokan \footnotemark[1]  \quad Lakshay Sharma \quad R. Venkatesh Babu \\
     Vision and AI Lab, Indian Institute of Science, Bangalore
 }

\begin{document}
\maketitle
\begin{abstract}
Vision-Language Models (VLMs) such as CLIP are trained on large amounts of image-text pairs, resulting in remarkable generalization across several data distributions. However, in several cases, their expensive training and data collection/curation costs do not justify the end application. This motivates a vendor-client paradigm, where a vendor trains a large-scale VLM and grants only input-output access to clients on a pay-per-query basis in a black-box setting. The client aims to minimize inference cost by \textit{distilling} the VLM to a student model using the limited available task-specific data, and further deploying this student model in the downstream application. While naive \textit{distillation} largely improves the In-Domain (ID) accuracy of the student, it fails to transfer the superior out-of-distribution (OOD) generalization of the VLM teacher using the limited available labeled images. To mitigate this, we propose Vision-Language to Vision - Align, Distill, Predict (VL2V-ADiP), which first aligns the vision and language modalities of the teacher model with the vision modality of a pre-trained student model, and further distills the aligned VLM representations to the student. This maximally retains the pre-trained features of the student, while also incorporating the rich representations of the VLM image encoder and the superior generalization of the text embeddings. The proposed approach achieves state-of-the-art results on the standard Domain Generalization benchmarks in a black-box teacher setting as well as a white-box setting where the weights of the VLM are accessible. Project page: \url{http://val.cds.iisc.ac.in/VL2V-ADiP/}
\end{abstract}    
\vspace{-0.5cm}
\section{Introduction}
\label{sec:intro}

\begin{figure}
\centering
    \includegraphics[width=0.9\linewidth]{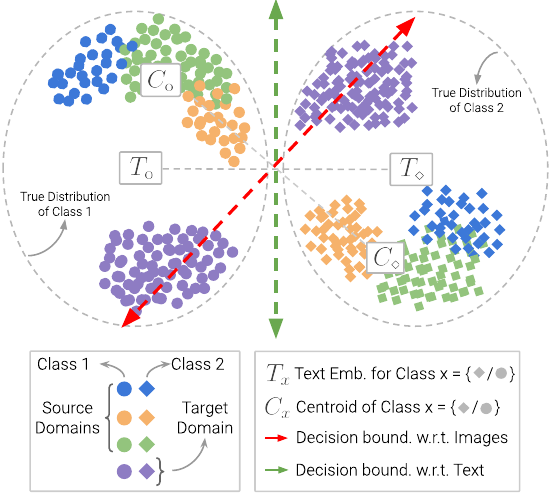}
    \vspace{-0.2cm}
    \caption{\textbf{Schematic diagram showing class and domain distributions in the shared text/ image embedding space of a VLM:} VLMs learn highly specialized image representations that are not domain invariant. Thus, a linear classifier (\textit{red decision boundary}) that is trained over the vision encoder using limited training data cannot generalize well to the target domain (\textit{shown in purple}). On the other hand, generic text embeddings such as ``\textit{A photo of a {class}}" represent the core concept of a class by virtue of their training method and vast training data. Thus, they generalize effectively across domains, and a zero-shot classifier (\textit{green decision boundary}) aligns better with the true distribution of classes.}
    \label{fig:teaser}
    \vspace{-0.5cm}
\end{figure}

While the initial success of Deep Learning was predominantly driven by training specialized models for each task or dataset \cite{lecun2015deep,krizhevsky2012imagenet}, recent research on foundation models \cite{CLIP,BLIP,ALIGN,LiT} eliminates the need for this by training generic models jointly over several modalities using large-scale data. The use of both image and language modalities in large-scale Vision-Language Models (VLMs) enables their use in several applications including zero-shot classification, where the embedding of an image is compared with text embeddings for ``\textit{a photo of a \{class\}}" corresponding to every class during inference and the class with highest similarity is predicted. For example, the LiT (private) model \cite{LiT} has been trained on 4 billion image-text pairs scraped from the web, and it achieves $85.2\%$ zero-shot accuracy on ImageNet. VLMs also demonstrate extraordinary performance across several distributions, owing to the vast diversity of distributions seen during their training \cite{CLIP}.

The remarkable OOD generalization of VLMs makes them suitable for use in several applications, either directly or after finetuning on the downstream target dataset. Although several open-sourced models such as CLIP \cite{CLIP} are easily accessible, they may not be suitable for use in specialized applications. For example, the training data may have to be curated to reduce biases in the model or to ensure data privacy. It may also be important to ensure that the web-crawled data is clean for adoption in critical applications such as autonomous driving, to avoid data-poisoning attacks \cite{biggio2012poisoning,carlini2023poisoning}. Further, general-purpose models cannot be used in specialized applications such as medical diagnosis, where a VLM trained on data containing expert comments can be invaluable. This motivates the need for training highly specialized VLMs, which can be expensive due to the high costs associated with data collection/ curation and training.

A given downstream application may not always justify the cost associated with training such large-scale models. However, since these VLMs are generic and applicable to several downstream applications, one vendor could train such a model, and make it available to several clients in a black-box setting on a pay-per-query basis. In this case, the client can minimize their inference time costs by distilling the VLM to a student model using the limited available data specific to the downstream task, and deploying the distilled model rather than the VLM. Since the key motivation of using VLMs is their generalization to several domains, reliable OOD performance of the distilled model is crucial.

In this work, we consider the problem of distilling a multimodal Vision-Language model to a unimodal Vision model using limited task-specific downstream data, targeted toward better OOD performance with respect to the training domains used for distillation. While standard distillation of VLMs would require training data comprising of images and their associated text, in this case, the training data consists of only labeled images, similar to a classification task. Moreover, since the superior generalization abilities of VLMs are a result of the large-scale data ($\sim400$ million in the case of CLIP) used for their training, it is challenging to transfer these generalization abilities using a limited dataset comprising around 10,000 (0.01 million) images \cite{OfficeHome,PACS,VLCS}. Lastly, since the VLM is considered to be a black box, the image encoder's weights cannot be used as initialization to the vision model for further fine-tuning. However, since the downstream task-specific data is limited and is insufficient to train models from scratch, we assume that the student is initialized with weights from a model pre-trained on a publicly available dataset such as ImageNet-1K \cite{krizhevsky2012imagenet}, as is common in a domain-generalization setting \cite{DomainBed}. 

While knowledge distillation \cite{hinton2015distilling} using the labeled task-specific images improves the In-Domain accuracy on the downstream task, the improvement in OOD accuracy is marginal due to the limited size of the dataset. To address this, we first analyze the robustness of the text and image embeddings from the VLM and highlight the importance of text embeddings for better OOD generalization (see Fig.\ref{fig:teaser}). Further, we propose \emph{VL2V-ADiP}: \emph{Vision-Language to Vision - Align, Distill, Predict}, to firstly align the features of the pre-trained student model with the text and image modalities of the VLM teacher, and further fine-tune the student backbone in a distillation framework. This improves the robustness of the student, while also aligning the student embeddings to the VLM's text embeddings corresponding to the respective class, thereby making the latter suitable for use as a classifier in the student without the need for further training. We summarize our contributions below: 

\begin{itemize}
    \item We investigate the robustness of image and text embeddings of VLMs, and highlight the importance of the text embeddings for better OOD generalization. 
    
    \item To demonstrate the superior generalization of text embeddings, we propose VL2V-SD - a self-distillation approach, that improves the OOD generalization of the VLM's pre-trained vision encoder using supervision from its own text encoder in a white-box VLM setting.

    \item We further propose VL2V-ADiP - a black-box distillation approach that effectively combines the features of a pre-trained vision model with the text and vision encoders of a VLM teacher for better OOD generalization.
    
    \item We demonstrate state-of-the-art results on standard Domain Generalization benchmark datasets in both white-box and black-box settings of the VLM teacher.
\end{itemize}

\section{Related Works}
\label{sec:rel_works}

\textbf{Vision-Language Models (VLMs):} VLMs are trained on large-scale datasets of image-text pairs crawled from the web \cite{YFCC100M, Conceptual12M, LAION}. Contrastive pre-training methods \cite{CLIP, ALIGN, LiT, BLIP}  maximize the similarity between the embeddings of matching image-text pairs. More recent efforts focus on scaling up image-text pre-training to larger datasets \cite{BLIP}, better supervision with unimodal and multimodal data \cite{FLAVA}, data-efficient pre-training \cite{DeCLIP}, and fine-tuning VLMs for downstream tasks \cite{WISE-FT, FLYP, LP-FT}. Another line of work focuses on leveraging CLIP \cite{CLIP}  to improve the OOD generalization in image classification \cite{CLIPood, text2concept} .

\noindent
\textbf{Domain Generalization (DG):} Prior works learn domain invariant representations by using augmentations \cite{nam2021reducing,nuriel2021permuted,robey2021model}, feature alignment across domains \cite{ganin2016domain,li2018adv,li2018domain,sun2016deep}, and disentangling domain-specific and domain-invariant features \cite{piratla2020efficient,lv2022causality,chattopadhyay2020learning}. Gulrajani \emph{et.al} \cite{DomainBed} demonstrate that a well-tuned ERM model is comparable to several past DG algorithms \cite{arjovsky2019invariant, sagawa2019distributionally, wang2020heterogeneous, li2018learning, 8578664, blanchard2021domain, zhang2021adaptive, krueger2021out, huang2020self}. SWAD \cite{SWAD} is a generic strategy that performs weight-averaging across different model snapshots in the optimal basin during training. MIRO \cite{MIRO} proposes Mutual Information (MI) regularization to maximally retain the pre-trained model's representations. SAGM \cite{SAGM} aligns gradient directions between the $\epsilon$-perturbed loss and the empirical risk. We demonstrate performance gains when compared to the existing DG methods in both white-box and black-box settings of the VLM. 

More recently, several methods perform diverse training of multiple models, followed by weight-averaging to improve OOD generalization \cite{DART, DiWA, EoA, Ratatouille}. However, they incur the enormous expense of training several diverse models. We thus restrict our comparison to DART \cite{DART}, where the number of models trained is lower than the others. Another direction has been the generation of better prompts for improved generalization of VLMs \cite{TPT, COOP, menon2022visual}. These methods can be integrated with the proposed approach.

\noindent\textbf{Knowledge Distillation (KD):} Knowledge distillation aims to transfer knowledge from a powerful teacher model to a compact student model. Output-distillation \cite{hinton2015distilling} uses the teacher's softmax outputs as guidance for the student. Feature-distillation methods \cite{FitNet, AlpKD, SemCKD, knowreview, SimKD, FCFD} use the teacher's intermediate representations as guidance since they contain richer information. The proposed method operates in the black box distillation setup that allows only input-output access to the teacher network and restricts access to intermediate features. We thus compare the proposed black-box approach only with methods that do not access intermediate features \cite{hinton2015distilling, SimKD, RISE}.

\begin{table}[]
\centering
\setlength{\tabcolsep}{3pt}
\caption{\textbf{CLIP on DG datasets:} Performance (\%) of CLIP ViT-B/16 model on using different embeddings in the classifier head for computing similarity w.r.t. the test image during inference (T.E.: Text Embedding, I.E.: Image Embedding).}
\label{tab:clip-zs}
\vspace{-2.5mm}
\resizebox{\linewidth}{!}{%
\begin{tabular}{@{}lccccc@{}}
\toprule
\rowcolor[HTML]{EFEFEF} 
\textbf{Embedding used for computing similarity} & \textbf{OH} & \textbf{TI} & \textbf{VLCS} & \textbf{PACS} & \textbf{Avg.} \\ \midrule
\begin{tabular}[c]{@{}l@{}}\textbf{E1:} Text embedding for "A photo of a \\ \{class\}"\end{tabular}                                    & 82.36       & 34.19       & 82.08         & 96.10         & 73.68         \\
\rowcolor[HTML]{EFEFEF} 
\begin{tabular}[c]{@{}l@{}}\textbf{E2:} Avg. text embedding for "A \{domain\} \\ of a \{class\}" across all train domains\end{tabular} & 83.70       & 35.55       & 82.28         & 96.21         & 74.44         \\
\begin{tabular}[c]{@{}l@{}}\textbf{E3:} Avg. image embedding of each class\\  (Source)\end{tabular}                                    & 71.37       & 33.99       & 48.21         & 79.03         & 58.15         \\
\rowcolor[HTML]{EFEFEF} 
\begin{tabular}[c]{@{}l@{}}\textbf{E4:} Avg. image embedding of each class\\  (Target)\end{tabular}                                    & 78.21       & 38.69       & 69.31         & 93.08         & 69.82         \\
\begin{tabular}[c]{@{}l@{}}\textbf{E5:} Avg. image embedding of 10 images \\ per class closest to test image (Source)\end{tabular}     & 76.42       & 39.33       & 76.42         & 92.15         & 71.08         \\
\rowcolor[HTML]{EFEFEF} 
\begin{tabular}[c]{@{}l@{}}\textbf{E6:} Avg. image embedding of 10 images \\ per class closest to test image (Target)\end{tabular}     & 84.86       & 85.38       & 87.88         & 98.32         & 89.11         \\ \bottomrule
\end{tabular}%
}
\vspace{-4mm}
\end{table}
\vspace{-0.2cm}
\section{Notations}
\label{sec:notation}

We consider the problem of Knowledge Distillation (KD) from VLMs (teacher) to vision models (student) for improved OOD generalization. We denote the VLM's text embedding for the input, \textit{``A photo of a \{class\}''} for class $c$ as $\mathbf{T}_{c}$. The image embedding corresponding to the image $x$ is denoted as $\mathbf{I}_x$. The student model's features corresponding to an input $x$ are denoted as $\textbf{F}_x$. These features when projected to the same dimension as the VLM embeddings ($\mathbf{I}_x$ and $\mathbf{T}_c$), are denoted as $\textbf{PF}_{x}$. The cosine similarity between two vectors $\textbf{a}$ and $\textbf{b}$ is denoted as $\cos(\textbf{a},\textbf{b})$. While we use CLIP \cite{CLIP} as the teacher for most of our analysis and experiments, we show the compatibility of the proposed approach with other VLMs \cite{FLAVA,BLIP,DeCLIP,yao2021filip} in Appendix \ref{sec:other_vlms}.
\section{Robustness of CLIP embeddings}
In this section, we present an overview of the training and zero-shot evaluation of the CLIP model, followed by a study on the robustness of its image and text embeddings.

\subsection{CLIP training and zero-shot prediction}
CLIP is a VLM that consists of an image encoder and a text encoder trained jointly on 400 million web-scraped image-text pairs \cite{CLIP}. The outputs of CLIP are the text and image embeddings for the respective text and image inputs. CLIP is trained using a contrastive loss which enforces similarity between the representations of positive image-text pairs and disparity between the representations of negative image-text pairs in the training minibatch. The training objective of CLIP is shown in Eq.\ref{eq:clip}, where the embeddings of the $i^{th}$ image-text pair are represented as $\mathbf{I}_i$ and $\mathbf{T}_i$ respectively.: 
\begin{equation}
\label{eq:clip}
        \mathcal{L_{\mathrm{CLIP}}}(x_i) = - ~ \textrm{log} \frac{\exp(\cos(\textbf{I}_i,\textbf{T}_i))}{\sum_{c=1}^{C} \exp(\cos(\textbf{I}_i,\textbf{T}_c))}
\end{equation}
The use of both text and image modalities during training enables the effective use of text embeddings as a zero-shot classifier, where inference can be performed directly without access to training data. For this, the text embeddings $\mathbf{T}_c$ corresponding to the captions \textit{``A photo of a \{class\}''} for each of the $C$ classes are compared with the image embedding $\mathbf{I}_i$, and the class with maximum similarity is predicted as shown in Eq.\ref{eq:pred}. Thus, the text embeddings $\mathbf{T}_c$ can be viewed as the weights of a classifier for a CLIP backbone.
\begin{equation}
\label{eq:pred}
    \hat{y}_i = \mathrm{argmax}_c \cos(\mathbf{I}_i, \mathbf{T}_c)
\end{equation}

\subsection{Characteristics of image and text embeddings}
\label{sec:embedding_robustness}
CLIP \cite{CLIP} shows remarkable zero-shot performance as shown in Table-\ref{tab:clip-zs} ($\mathrm{E1}$), and outperforms even an ImageNet pre-trained model that is explicitly fine-tuned using the downstream dataset (Table-\ref{tab:main-dg}) on several DG datasets. To understand its superior zero-shot performance, we discuss below the robustness of its image and text embeddings.

\textbf{Characteristics of the Image Encoder:}
In image classification tasks, the representations learned using standard ERM training are expected to exhibit invariances to several factors of variation due to: (a) the classification objective that enforces similar representations for different variations within a given class, which are different from the representations for other classes and (b) augmentations such as color jitter that enforce additional invariance. However, CLIP is trained using detailed captions for each image, such as \emph{``A brown cat sitting on a sofa''}, \emph{``A black cat standing on two limbs''}, and \emph{``A black dog with long ears and a lot of fur''}. This allows CLIP to learn rich specialized representations for each attribute, with higher intra-class variance when compared to standard ERM training.

\textbf{Characteristics of the Text Encoder:} Robustness to distribution shifts can be achieved either by incorporating the required domain-invariance in the feature extractor, or by domain-invariant feature selection at the classification head. As discussed, CLIP's image encoder produces input-specific representations and is not domain invariant. The superior generalization of CLIP on different datasets and domains is thus a result of the robust text embeddings. Although descriptive captions such as ``A brown cat sitting on a sofa'' are expected to have unique embeddings based on pose, location, etc., generic captions such as ``A photo of a \{class\}'' represent the core concept of the class.

\textbf{Empirical observations:} In Table-\ref{tab:clip-zs}, we present results to demonstrate the robustness of CLIP text embeddings on the multi-source Domain Generalization benchmarks, where each dataset consists of images from $n$ domains. The results presented are an average of $n$ cases, where each case corresponds to the training on a given set of $n-1$ domains and testing on the remaining unseen domain referred to as the target domain. Firstly, CLIP zero-shot results ($\mathrm{E}1$) show the robustness of text embeddings across different distributions. The robustness of the text embeddings can be improved by explicitly enforcing domain invariance for better generalization to unseen domains ($\mathrm{E}2$). However, the same level of robustness is not seen when we use an average of all image embeddings across the same domains for a given class, as shown in $\mathrm{E}3$. This is due to the fact that while the text embeddings for ``A \{domain\} of a \{class\}'' are obtained by training over a large dataset (400 million images), the image embeddings are an average of $\sim0.01$ million images on the downstream dataset. This is illustrated in Fig.\ref{fig:teaser}, where the red decision boundary trained using image embeddings misclassifies images from the unseen target domain, while the green classifier trained using the generic text embeddings is closer to an ideal decision boundary. However, an average image embedding on the target domain improves results significantly ($\mathrm{E}4$). The accuracy improves further when an average embedding corresponding to a few (10) source domain images closest (in terms of cosine similarity of image embeddings) to the target image is used for each class during prediction ($\mathrm{E}5$). A similar experiment using images from the target domain ($\mathrm{E}6$), achieves considerable improvements over the zero-shot accuracy in $\mathrm{E}1$. We can consider $\mathrm{E}6$ as an In-(target)-Domain setting where a labeled hold-out validation set is used during prediction for obtaining the average embeddings. Therefore, it is possible to outperform the generalized representations of the text embeddings only when labeled images from the target domain are available. \emph{Thus, in a DG setting where the target domain is inaccessible, the generic text embeddings provide the best robustness across distribution shifts}. We therefore explicitly use the text embeddings to maximally transfer their robustness to the student model. 
\section{Proposed Approach: VL2V-ADiP}

In this section, we first present the knowledge distillation framework adapted to a VLM-to-vision distillation setting. We further propose a self-distillation approach VL2V-SD, to enhance the robustness of the VLM image encoder, which motivates the proposed approach VL2V-ADiP.

\subsection{Distillation from VLMs to Vision models}
\label{sec:kd}
The standard framework of Knowledge Distillation (KD) \cite{hinton2015distilling} applies in the unimodal setting, for instance, where the teacher and student are both vision models. To adapt this to a VLM-to-vision distillation setting, a classification model (which we refer to as a VLM-classifier) is first constructed using the VLM. The feature extractor of the VLM-classifier comes from the image encoder of the VLM, and the linear classifier head obtains its weights from the text embeddings $\textbf{T}_c$ corresponding to the captions ``a photo of a \{class\}'' for all $C$ classes. For every image-label pair $(x_i,y_i)$, the VLM-classifier first computes the image embedding $\textbf{I}_{x_i}$, and further computes its similarity w.r.t. each of the text embeddings $\textbf{T}_c$. The similarity scores $\cos(\textbf{I}_{x_i}, \textbf{T}_c)$ are considered to be analogous to the logit layer, over which the softmax function is applied. The final distillation loss of the student that is minimized during training is shown below in Eq.\ref{eq:KD} for a minibatch of size $n$. Here, $f_S(.)$ and $f_\mathrm{T}(.)$ represent the softmax representations of the student and teacher respectively, $\mathrm{CE}$ represents the Cross-Entropy loss, and $\mathrm{KL}$ represents the Kullback-Leibler divergence. Temperature scaling is applied to the logits before computing softmax \cite{hinton2015distilling}. We tune over the range - \{0.1, 1, 10\} on the in-domain validation set, and report the baseline best results in Table-\ref{tab:main-dg}.
\begin{equation}
\label{eq:softmax}
    f_\mathrm{T}(x_i) = \mathrm{softmax}(\cos(\textbf{I}_{x_i}, \textbf{T}_c))
\end{equation}
\begin{equation}
\label{eq:KD}
    \mathcal{L}_\mathrm{KD} = \frac{1}{n} \sum_{i=1}^n \big\{\mathrm{CE}(f_S(x_i), y_i) + \lambda \cdot \mathrm{KL}\big(f_\mathrm{T}(x_i) || f_S(x_i)\big)\big\}
\end{equation}

\begin{figure*}[t]
\centering
    \includegraphics[width=1.0\linewidth]{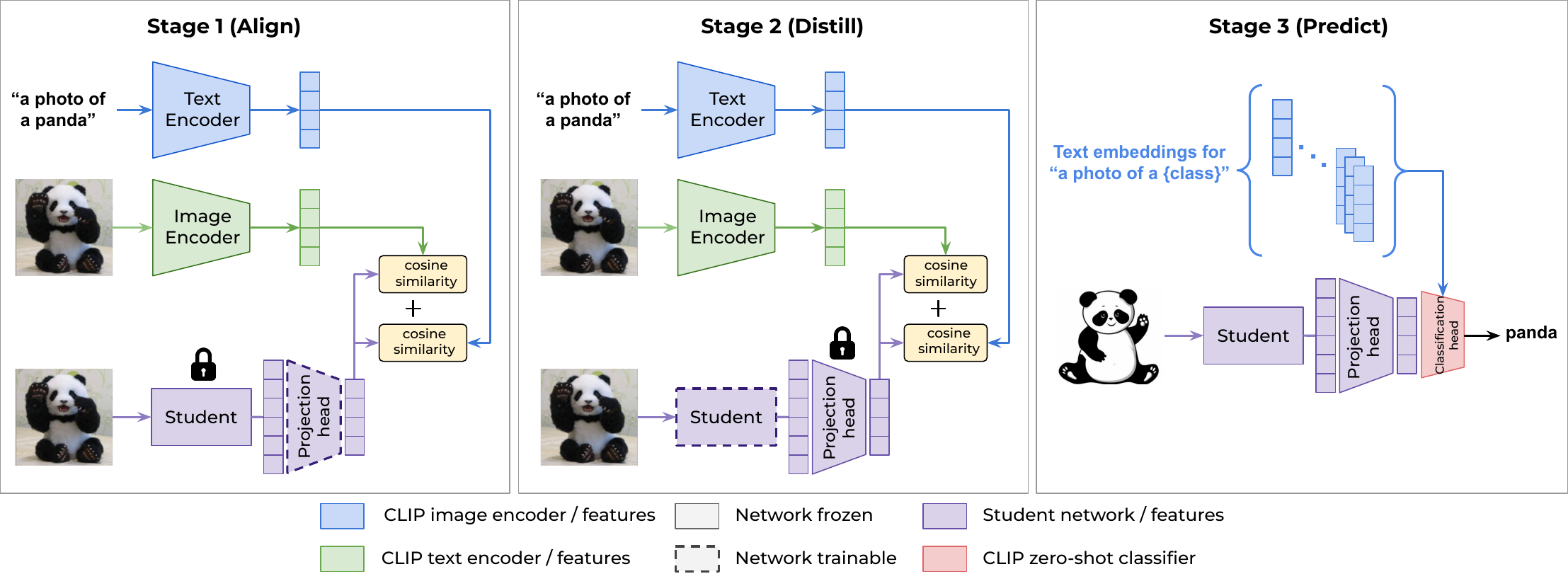}
    \caption{Overview of the proposed approach VL2V-ADiP, consisting of (a) Align, (b) Distill and (c) Predict Stages for Black-Box Distillation from Vision-Language to Vision (VL2V) models.}
    \label{fig:main_fig}
    \vspace{-0.3cm}
\end{figure*}

\subsection{Self-Distillation from Text to Image encoders}
\label{sec:self-distillation}
As discussed in Section-\ref{sec:embedding_robustness}, the image encoder of CLIP is trained to represent all factors of variations in an image that are described in the corresponding caption used during its training. Therefore, it does not exhibit the invariances to factors such as color, texture, lighting, and pose, that are required for an image classification task. On the other hand, text prompts allow much better control over invariances in text-embeddings. For example, a prompt such as ``A brown cat sitting on a sofa'' uniquely represents the color, location, and pose of the cat in addition to the object itself, whereas a prompt ``A photo of a cat'' is invariant to such variations.

Motivated by this, we propose \emph{Vision-Language to Vision Self Distillation (VL2V-SD)}, where the invariances of generic text embeddings are distilled to the image encoder using the downstream dataset. We minimize the loss shown in Eq.\ref{eq:vl2v-sd} for training, where $\textbf{T}_{y_i}$ represents the embedding for ``A photo of a \{class\}'' corresponding to the ground truth class $y_i$, and $\textbf{I}^t_{x_i}$, $\textbf{I}^s_{x_i}$ represent the image embeddings of the teacher and student respectively for an image $x_i$. 

\vspace{-5mm}
\begin{equation}
    \label{eq:vl2v-sd}
    \mathcal{L}_{\mathrm{SD}} = - \frac{1}{2n} \sum_{i=1}^n \big\{\cos(\textbf{I}^s_{x_i}, \textbf{T}_{y_i}) + \cos(\textbf{I}^s_{x_i}, \textbf{I}^t_{x_i})\big\}
    \vspace{-0.1cm}
\end{equation}

The image encoders of both teacher and student are the same (VLM's image encoder) initially, and the student's image encoder is updated during training. The text encoder is frozen during training. While the first term enforces the image encoder to learn representations that match the corresponding text embeddings more accurately on the given dataset, the second term ensures that there is no representation collapse, and retains the rich features learned by the image encoder. The classifier is composed of text embeddings corresponding to the prompts ``a photo of a \{class\}'' for all classes, as was the case in the VLM-classifier discussed in Section-\ref{sec:kd}. We present the results of VL2V-SD in Table-\ref{tab:sd}, where we note significant improvements over both CLIP zero-shot and the SOTA DG methods - SWAD \cite{SWAD}, MIRO \cite{MIRO}, DART \cite{DART}, and SAGM \cite{SAGM} on the ViT-B/16 architecture with CLIP initialization. We demonstrate significant gains over the CLIP fine-tuning methods LP-FT \cite{LP-FT}, FLYP \cite{FLYP},  CLIPood \cite{CLIPood}, and WiSE-FT \cite{WISE-FT}. We also achieve significant gains over a distillation method RISE \cite{RISE}, despite using the standard text prompt while RISE uses an expanded set of prompts. For WiSE-FT, we present the results with the optimal $\alpha$ that achieves the best performance by using the search space $\alpha=[0.1-0.9]$. Similarly, for RISE, we present the results with the optimal hyperparameters $\lambda_1, \lambda_2, \lambda_3$ based on their search space. We incorporate SWAD \cite{SWAD} in all our runs for the baselines as well as the proposed approach (denoted by ``(S)") since it is a generic technique that improves the generalization of any given method. However, we do not incorporate SWAD with WiSE-FT since it already performs weight averaging of the pre-trained and fine-tuned models. We present the results without SWAD in Appendix \ref{sec:swad_comparisons}.

\begin{table}[]
\centering
\setlength{\tabcolsep}{5pt}
\caption{\textbf{White-Box setting (CLIP initialization):} Performance (\%) of the proposed approach VLV2-SD, compared to the existing methods. ViT-B/16 architecture is used. (S) denotes SWAD \cite{SWAD}.}
\vspace{-2mm}
\label{tab:sd}
\resizebox{\linewidth}{!}{%
\begin{tabular}{l|ccccc|cc}
\toprule
  \textbf{Method} &
  \textbf{OH} &
  \textbf{TI} &
  \textbf{VLCS} &
  \textbf{PACS} &
  \textbf{DN} &
  \textbf{Avg-ID} &
  \textbf{Avg-OOD} \\ \midrule
Zero-shot \cite{CLIP}   & 82.40 & 34.10          & 82.30 & 96.50 & 57.70 & -              & 70.60 \\
SWAD \cite{SWAD}        & 81.01 & 42.92          & 79.13 & 91.35 & 57.92 & 89.05          & 70.47 \\
MIRO (S) \cite{MIRO}        & 84.80 & \textbf{59.30} & 82.30 & 96.44 & 60.47 & \textbf{91.00} & 76.66 \\
DART (S) \cite{DART}        & 80.93 & 51.24          & 80.38 & 93.43 & 59.32 & 89.25          & 73.06 \\
SAGM (S) \cite{SAGM}        & 83.40 & 58.64          & 82.05 & 94.31 & 59.05 & 89.74          & 75.49 \\
LP-FT (S) \cite{LP-FT}      & 81.17 & 47.26          & 80.88 & 92.92 & 57.04 & 88.97          & 71.85 \\
FLYP (S) \cite{FLYP}       & 82.76 & 33.25          & 66.64 & 78.53 & 57.41 & 78.94          & 63.72 \\
CLIPood (S) \cite{CLIPood}  & 83.31 & 46.28          & 77.19 & 93.16 & 57.78 & 69.90          & 71.55 \\
WiSE-FT (S) \cite{WISE-FT} & 86.32 & 54.50 & 82.88 & \textbf{97.29} & 58.01 & 88.35 & 75.80 \\
 RISE (S) \cite{RISE}& 78.39& 49.61& 80.62& 93.25& 55.37& 87.91&71.45\\
\textbf{VL2V-SD (Ours)} &
  \textbf{87.38} &
  58.54 &
  \textbf{83.25} &
  96.68 &
  \textbf{62.79} &
  89.99 &
  \textbf{77.73} \\ \bottomrule
\end{tabular}%
}

\vspace{-4mm}
\end{table}

\subsection{VL2V - Align, Distill, Predict (VL2V-ADiP)}
\label{sec:VL2V-ADiP}
While VL2V-SD is very effective in improving the performance of the VLM in a white-box setting, it does not address the problem of \emph{black-box distillation} that we consider in this work, since it assumes access to the weights of the VLM model. In this section, we discuss how this method can be adapted to the black-box setting and present the proposed approach \emph{VL2V-ADiP}. Since the amount of downstream task-specific data is assumed to be limited, the student is initialized with the best available pre-trained model such as ImageNet \cite{krizhevsky2012imagenet}.

In the self-distillation case seen in Section-\ref{sec:self-distillation}, the goal was to induce domain invariance from the text encoder to the image encoder, while ensuring that the rich features learned by the image encoder are not lost. Whereas, in the black-box setting, the goal is to distill the rich features learned by the image-encoder of the VLM and the domain invariance from the text embeddings. While the two cases can have a similar loss formulation as shown in Eq.\ref{eq:vl2v-sd}, the method VL2V-SD cannot directly be applied here since the feature dimension of the student model may be different from the dimension of VLM's image and text embeddings. Even if the dimensions had matched, the features of the ImageNet pre-trained backbone and VLM's embeddings would be misaligned. Directly enforcing a similarity loss in such a case would lead to forgetting of the pre-trained features. To address this, we propose \emph{VL2V-ADiP, Vision-Language to Vision - Align, Distill, Predict} as shown in Fig.\ref{fig:main_fig} (Ref: Alg. \ref{algo:overall} in the Appendix). We propose to first \emph{Align} the features of the pre-trained student model with VLM embeddings using a linear projection head on the student backbone in Stage-1, as shown in Fig.\ref{fig:main_fig}(a). For this, the student backbone is frozen and only the projection head is trained using the following loss, where $\textbf{PF}^s_{x_i}$ represents the projected features of the student model for the input $x_i$:
\begin{equation}
\label{eq:vl2v-adip}
    \mathcal{L}_{\mathrm{ADiP}} = - \frac{1}{2n} \sum_{i=1}^n \big\{\cos(\textbf{PF}^s_{x_i}, \textbf{T}_{y_i}) + \cos(\textbf{PF}^s_{x_i}, \textbf{I}^t_{x_i})\big\}
\end{equation}
Next, the aligned student features are refined using the VLM's image and text embeddings in the \emph{Distill} step as shown in Fig.\ref{fig:main_fig}(b) (Stage-2). For this step, the linear projection head is frozen, and the feature extractor is trained using the same loss as Stage-1, as shown in Eq.\ref{eq:vl2v-adip}. While we give equal weight to both loss terms in Eq.\ref{eq:vl2v-adip}, we present the impact of varying this in Appendix \ref{sec:loss_weight}. Finally, in Stage-3 (Predict), the VLM's embeddings corresponding to ``a photo of a \{class\}'' for all $C$ classes are used as weights of the classification head, and the class with the highest similarity to the given image embedding is predicted, as shown in Eq.\ref{eq:pred}. In addition to learning the rich features from the VLM teacher's image-encoder and invariances from its text encoder, enforcing similarity between the features of the student backbone and the text embeddings of the corresponding classes also makes the text embeddings more suitable for use as a classifier in Stage-3 (Predict). Thus, although Stages 1 and 2 explicitly train only the student backbone and projection head, they implicitly impact the effectiveness of the classification head as well. This facilitates the use of these embeddings directly as a classifier without the need for fine-tuning further on the downstream dataset. As shown in Table-\ref{tab:method_abl}, further fine-tuning degrades the performance, since only In-Domain (ID) data is available for fine-tuning, which would make the classifier forget the domain invariance that it inherently possesses. However, fine-tuning with limited In-Domain data improves the ID accuracy as shown in the table.
\section{Experiments and Results}
\label{sec:experiment}

\subsection{Evaluation Details} 
In this work, we consider distillation from VLMs to Vision models for improving the OOD generalization of the latter on vision tasks. We, therefore, consider the five popular Domain Generalization (DG) datasets for the empirical evaluation - OfficeHome (OH) \cite{OfficeHome}, PACS \cite{PACS}, VLCS \cite{VLCS}, Terra-Incognita (TI) \cite{TerraIncognita} and DomainNet (DN) \cite{DomainNet}. Each of the datasets consists of $d$ domains, where $d=4$ for the first four datasets and $d=6$ for DomainNet. We present more details on the datasets in Appendix \ref{sec:datasets}. We follow the DomainBed \cite{DomainBed} framework for training and evaluation, where each domain is split into training and validation sets, and training is performed on $d-1$ domains while evaluation is done on the $d^{th}$ unseen domain. Thus, for every dataset, $d$ models are trained, by leaving one domain out for evaluation each time. The average accuracy across these $d$ runs on the unseen test domain (OOD accuracy) and the average In-domain (ID) validation split accuracy are reported. The unseen test domain is neither used for training nor for hyperparameter tuning \citep{DomainBed}. 

\subsection{Training Details} 
The number of training iterations is set to $5000$ for OfficeHome, PACS, VLCS, and Terra-Incognita, and to $15000$ for DomainNet, as is standard \cite{SWAD, MIRO}. We use the Adam optimizer with a constant learning rate of $5 \times 10^{-5}$ for all our experiments except CLIPood (S) in Table-\ref{tab:sd}, for which we use the AdamW optimizer as specified in \cite{CLIPood}. These settings are fixed for all datasets, and we do not introduce any additional hyperparameters in the proposed algorithm. Our primary evaluations consider a ViT-B/16 architecture \cite{dosovitskiy2020image} for both the VLM (teacher) Image-encoder and the student model to enable a fair comparison with existing methods, that use the same architecture. Since one of the key motivations of distillation is to train a low-capacity model, we present results with lower-capacity student architectures as well. We use a CLIP teacher model, which uses a Transformer architecture \cite{vaswani2017attention} with modifications described by \citet{radford2019language} for the text encoder. The student model and our primary baselines for VL2V-ADiP use an ImageNet pre-trained initialization \cite{krizhevsky2012imagenet}, as is standard in DG \cite{DomainBed}. We obtain the results of baseline methods on ViTs in Tables \ref{tab:sd} and \ref{tab:main-dg} by integrating the official code of the respective method with the MIRO \cite{MIRO} code base to ensure a fair comparison. Additionally, we show results with other teacher VLMs in Appendix \ref{sec:other_vlms}. 

\begin{table}[t]
\centering
    \caption{\textbf{SOTA comparison with ImageNet initialization:} Performance (\%) of the proposed approach VL2V-ADiP, compared to existing KD and DG methods (with SWAD). ViT-B/16 with ImageNet-1K initialization is used as the student.} 
    \label{tab:main-dg}
    \vspace{-3mm}
\resizebox{\linewidth}{!}{%
\begin{tabular}{@{}l|ccccc|cc@{}}
\toprule
  \textbf{Method} &
  \textbf{OH} &
  \textbf{TI} &
  \textbf{VLCS} &
  \textbf{PACS} &
  \textbf{DN} &
  \textbf{Avg-ID} &
  \textbf{Avg-OOD} \\ \midrule

ERM-LP       & 71.48 & 31.35 & 77.52 & 67.02 & 36.65 & 73.99 & 56.81 \\
ERM Full Fine-tuning      & 83.22 & 50.05 & 80.33 & 90.28 & 56.10 & 89.31 & 72.00 \\
LP-FT \cite{LP-FT}  & 81.55	& 51.61	& 80.17	& 91.20 & 56.03 & \textbf{90.03} & 72.11 \\
SimKD \cite{SimKD} & 66.76 & 81.01 & \textbf{83.92}& 28.24 & 49.42 & 68.24 & 61.87 \\
KD    \cite{hinton2015distilling} & 82.73 & 48.40 & 80.48 & 91.46 & 56.11 & 89.20 & 71.84 \\
MIRO \cite{MIRO} & 80.09 & 50.29 & 81.10 & 89.50 & 55.75 & 88.71 & 71.35 \\
DART \cite{DART} & 83.75 & 49.68 & 77.29 & 90.55 & 58.05 & 88.54 & 71.86 \\
SAGM \cite{SAGM} & 82.22 & 53.24 & 79.60 & 90.02 & 55.66 & 89.22 & 72.15 \\
Text2Concept \cite{text2concept} & 70.24 & 26.46 & 64.77 & 79.03 & 23.26 & 53.15 & 52.82 \\
RISE \cite{RISE} & 83.48 & 52.55 & 83.70 & 93.54 & 56.58 & 88.91 & 73.97 \\
\textbf{VL2V-ADiP (Ours)} &
  \textbf{85.74} &
  \textbf{55.43} &
  81.90&
  \textbf{94.94} &
  \textbf{59.38} &
  89.02 &
  \textbf{75.48} \\ \bottomrule
\end{tabular}%
}
\end{table}
\begin{table}[t]
\centering
\setlength{\tabcolsep}{10pt}
\caption{\textbf{Distillation to lower capacity student models:} Performance (\%) of the proposed approach VL2V-ADiP (denoted as Ours), when compared to the KD baseline \cite{hinton2015distilling} and ERM (S) or SWAD \cite{SWAD} with different student architectures. The teacher architecture is ViT-B/16. (S) denotes SWAD.}
\vspace{-3mm}
\label{tab:other-students}
\setlength{\tabcolsep}{5pt}
\resizebox{1.0\linewidth}{!}{%
\begin{tabular}{@{}cl|ccccc|cc@{}}

\toprule
\textbf{Student} & \textbf{Method} & \textbf{OH} & \textbf{TI} & \textbf{VLCS} & \textbf{PACS} & \textbf{DN} & \textbf{Avg-ID} & \textbf{Avg-OOD} \\ \midrule

\multirow{3}{*}{\begin{tabular}[c]{@{}c@{}}ViT-B/16 \\ (86M)\end{tabular}}& ERM (S) & 83.22 & 50.05 & 80.33 & 90.28 & 56.10 & \textbf{89.31}& 72.00 \\
 & KD (S) & 82.73 & 48.40 & 80.48 & 91.46 & 56.17 & 89.20& 71.85 \\
 & Ours (S)& \textbf{85.74} & \textbf{55.43} & \textbf{81.90} & \textbf{94.94} & \textbf{59.38} & 89.02& \textbf{75.48} \\ \midrule

\multirow{3}{*}{\begin{tabular}[c]{@{}c@{}}ViT-S/16 \\ (22M)\end{tabular}} & ERM (S) & 78.58 & 49.40 & 78.72 & 85.80 & 52.09 & 85.66& 68.92 \\
 & KD (S) & 78.14 & 50.11 & 79.14 & 85.97 & 52.05 & \textbf{87.43}& 69.08 \\
 & Ours (S)& \textbf{81.22} & \textbf{52.47} & \textbf{81.44} & \textbf{89.32} & \textbf{54.23} & 86.59& \textbf{71.73} \\ \midrule

\multirow{3}{*}{\begin{tabular}[c]{@{}c@{}}DeiT-S/16 \\ (22M)\end{tabular}} & ERM (S) & 74.95 & 47.85 & 79.37 & \textbf{89.22} & 49.18 & 86.58& 68.12 \\
 & KD (S) & 74.65 & 48.11 & 78.86 & 88.14 & 49.10 & \textbf{86.66}& 67.77 \\
 & Ours (S)& \textbf{77.63} & \textbf{48.72} & \textbf{81.89} & 88.97 & \textbf{50.37} & 85.28& \textbf{69.52} \\ \midrule

\multirow{3}{*}{\begin{tabular}[c]{@{}c@{}}ResNet-50 \\ (26M)\end{tabular}} & ERM (S) & 70.85 & 49.47 & \textbf{79.50} & \textbf{88.05} & 46.43 & 84.88& 66.86 \\
 & KD (S) & 70.67 & 51.22 & 78.63 & 87.23 & 46.31 & \textbf{85.04}& 66.81 \\
 & Ours (S)& \textbf{74.42} & \textbf{53.46} & 79.23 & 86.72 & \textbf{47.74} & 84.73& \textbf{68.31} \\ \bottomrule
\end{tabular}%
}
\vspace{-0.5cm}
\end{table}
\begin{table*}[ht]
\centering
\setlength{\tabcolsep}{8pt}
\caption{\textbf{Ablation Study of the proposed VL2V-ADiP:} Performance (\%) on ViT-B/16 model with ImageNet initialization.}
\vspace{-0.2cm}
\label{tab:method_abl}
\resizebox{\textwidth}{!}{
\begin{tabular}{lcccccc}

\toprule
\multicolumn{1}{c}{\cellcolor[HTML]{FFFFFF}\textbf{Method - Changes done w.r.t. VL2V-ADiP (Ours)}} & \textbf{OH} & \textbf{~~TI~~} & \textbf{VLCS} & \textbf{PACS} & \textbf{\begin{tabular}[c]{@{}c@{}}Avg-ID\end{tabular}} & \textbf{\begin{tabular}[c]{@{}c@{}}Avg-OOD\end{tabular}} \\
\midrule

\textbf{VL2V-ADiP (Ours)} & 85.74 & 55.43 & 81.90 & 94.94 & 92.74 & 79.50 \\
A1: Combining "Align" and "Distill" stages & 74.55 & 52.99 & 80.14 & 85.29 & 90.55 & 73.24 \\
A2: Without freezing projection head in Stage-2 & 86.13 & 56.68 & 81.78 & 93.75 & 93.00 & 79.59 \\
A3: Distilling only from Text encoder in Stages-1 and 2 & 83.18 & 47.02 & 79.83 & 90.86 & 91.87 & 75.22 \\
A4: Distilling only from Image encoder in Stages-1 and 2 & 78.97 & 28.97 & 82.24 & 89.97 & 74.09 & 70.04 \\
A5: Finetuning CLIP classifier-head in Stage-3 (CE loss) - CLIP init classifier & 84.47 & 49.28 & 81.30 & 93.54 & 93.06 & 77.15 \\
A6: Finetuning full network in Stage-3 (CE loss) -  CLIP init classifier & 83.88 & 49.83 & 80.13 & 92.25 & 93.37 & 76.52 \\
A7: Finetuning classifier-head in Stage-3 (CE loss) - random init classifier & 84.63 & 49.55 & 81.29 & 93.57 & 93.02 & 77.26 \\
A8: Finetuning full network in Stage-3 (CE loss) - random init classifier & 83.16 & 50.03 & 79.78 & 92.15 & 93.14 & 76.28 \\
\bottomrule
\end{tabular}}
\vspace{-0.1cm}
\end{table*}

\subsection{Comparison with the SOTA} 

We present the ID and OOD results of the proposed approach VL2V-ADiP when compared to the DG and KD baselines in Table-\ref{tab:main-dg}. \citet{DomainBed} show that ERM training is a very strong baseline in DG and outperforms several older methods that were proposed. Thus, ERM training of only the linear layer (ERM-LP) and ERM full fine-tuning (ERM-FFT) are two important baselines we consider. Further, we present the results of important SOTA DG methods - SWAD \cite{SWAD}, MIRO \cite{MIRO}, DART \cite{DART}, and SAGM \cite{SAGM}. Additionally, we also show results with Text2Concept \cite{text2concept} and LP-FT \cite{LP-FT} - simple fine-tuning strategies for improved OOD generalization on downstream tasks. While there are several methods that outperform ERM \cite{nam2021reducing,kim2021selfreg,sun2016deep,bui2021exploiting}, we do not present all of them here since the above-listed recent methods outperform them. Since we incorporate SWAD in the proposed approach, we present all the baseline results by integrating SWAD with each of them and denote them with a suffix ``(S)" in the tables. Additionally, we also present three distillation results - (a) knowledge distillation from the VLM teacher to the vision student model as discussed in Section-\ref{sec:kd} and (b) distillation from the VLM teacher using SimKD \cite{SimKD}, and (c) distillation from the VLM teacher using RISE \cite{RISE}. Since the feature distillation methods listed in Sec. \ref{sec:rel_works} cannot be extended to the black box distillation setting, we do not compare with them. The proposed approach VL2V-ADiP achieves $\sim$3.3\% improvement on average OOD accuracy across all datasets with respect to the best baselines, with comparable ID accuracy. We note that Text2Concept scales poorly on smaller datasets compared to the results shown in the paper \cite{text2concept}. SimKD \cite{SimKD} is a method intended for unimodal distillation and thus fails for the multimodal  setup. Furthermore, we outperform the method RISE \cite{RISE} by a large margin despite the latter using expanded text prompts. This shows that the proposed distillation method VL2V-ADiP effectively distills from VLMs with lesser prompt engineering and text supervision compared to RISE.
\begin{figure*}
\centering
        \includegraphics[width=1\linewidth]{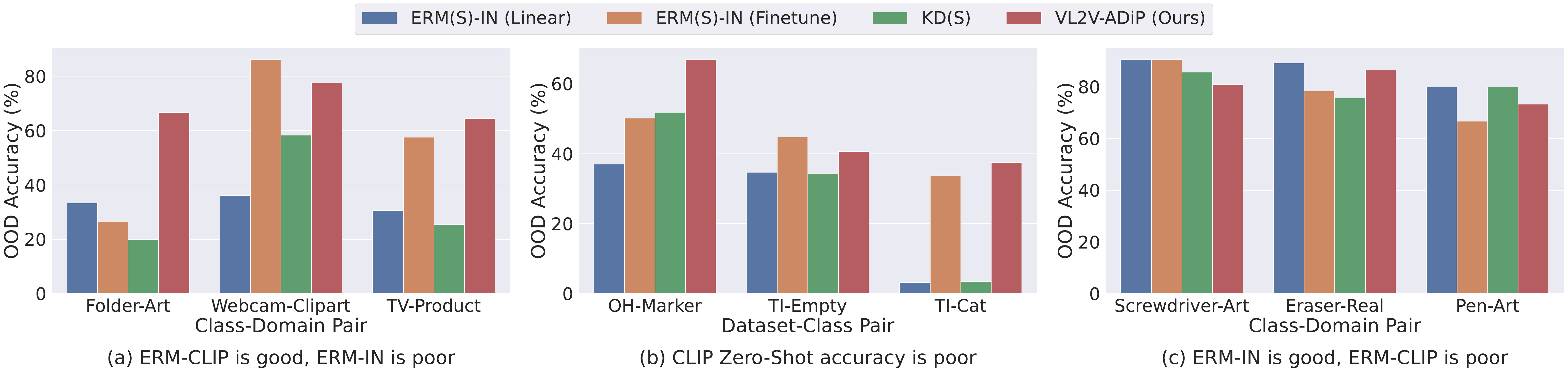}
        \vspace{-0.5cm}
    \caption{OOD accuracy (\%) of the proposed approach when compared to KD and ERM baselines for select classes/ domains in OfficeHome and Terra-Incognita datasets, where the (a) ERM-IN (linear) (b) CLIP Zero-shot (c) ERM-CLIP (linear) performance is poor.}
    \label{fig:bar-plot}
    \vspace{-0.5cm}
\end{figure*}

\subsection{Distillation to lower capacity student models} 
We present results of distilling from a CLIP ViT-B/16 teacher to different student architectures and capacities (ViT-S/16, DeiT-S, ResNet-50) in Table-\ref{tab:other-students}. We observe substantial gains over the Knowledge Distillation and ERM (S) or SWAD baseline across all student architectures. We note that the distillation performance is best when the architecture (ViT) of the teacher matches that of the student. For example, although ViT-S/16, DeiT-S/16, and ResNet-50 have a similar number of parameters, the performance of the ViT-S/16 student is the best, showing the architecture plays a role in the nature of representations learned, thereby allowing better transfer when they match. 

\subsection{Ablation Study}
We present an ablation study of the proposed method in Table-\ref{tab:method_abl}. Firstly, we study the impact of combining stages-1 (Align) and 2 (Distill), i.e., training the linear projection layer and the backbone jointly in $\mathrm{A}1$. This results in a 2.2\% drop in ID accuracy and a 6.3\% drop in OOD accuracy, highlighting that the pre-trained ImageNet features are retained better when they are aligned before distillation. In $\mathrm{A}2$, the linear projection head is also trained in Stage-2. The results of this case are similar to the proposed approach, showing that even when the linear head is changed, it does not get updated much after Stage-1 (Align). We next analyze the importance of distilling from both image and text encoders of CLIP by using only one of them individually in $\mathrm{A}3$ and $\mathrm{A}4$. In both cases, we note a significant drop in the OOD accuracy highlighting the importance of distilling from both encoders. Distilling only from the Image encoder has a larger drop due to the lack of alignment between the representations and the classifier head which is composed of CLIP text embeddings corresponding to each of the classes.

In Stage-3 (Predict) of the proposed approach, the CLIP text embeddings corresponding to the classes are directly used as the classifier weights. We explore the impact of training this classifier further using cross-entropy loss over the softmax representations shown in Eq.\ref{eq:softmax}, in experiments $\mathrm{A}5$-$\mathrm{A}8$. In $\mathrm{A}5$ and $\mathrm{A}6$, the classifier head is first initialized using the text embeddings of CLIP, and in $\mathrm{A}7$ and $\mathrm{A}8$, a random initialization is used for the classification head. In $\mathrm{A}5$ and $\mathrm{A}7$, only the classifier head is trained, whereas in $\mathrm{A}6$ and $\mathrm{A}8$, the full network is trained. All these experiments result in a drop in the OOD accuracy, indicating that training on the downstream dataset using the cross-entropy loss can destroy the domain invariances learned from CLIP. Since the impact of training the full network is higher, the drop is higher in this case when compared to finetuning only the classification head. In all cases, the In-Domain accuracy improves when compared to the proposed approach, with higher gains in the case when the full network is trained, confirming that training without the supervision of the CLIP teacher indeed overfits to the training domains. \par

We also compare the ERM and KD baselines with the proposed approach in some of the extreme cases in Fig.\ref{fig:bar-plot}, where the CLIP Image encoder is significantly better than the ERM model (a), and vice-versa (c). For this, we select the classes/domains with the highest difference (40-50\%) between the ERM-CLIP (Linear) and ERM-ImageNet (IN) (Linear) baselines from the OfficeHome dataset. We also present the case where the CLIP zero-shot accuracy is poor in OfficeHome (22\%) and Terra-Incognita datasets (0\%) in Fig.\ref{fig:bar-plot} (b). The KD baseline relies a lot on the ImageNet pre-trained backbone, hence it is poor when ERM-IN is poor, and is close to our approach when ERM-IN is good. The ERM fine-tuning baseline is best when the ImageNet features are much better than CLIP Image encoder's features, but it is poor in the other case, as expected. Lastly, even when CLIP zero-shot accuracy is poor (close to 0\%), the proposed approach achieves reasonable accuracy, since Stages-1 and 2 align the text embeddings better to the feature extractor, making them more suitable as a classifier. 
\vspace{-0.1cm}
\section{Conclusion}
\label{sec:conclusion}

In this work, we aim to leverage the superior generalization of large-scale Vision-Language Models towards improving the OOD generalization of Vision models. We consider a practical scenario where a client gets only black-box access to the model on a pay-per-query basis from the vendor, motivating the need for distilling the model first, and further using the student model during inference. Towards this, we first highlight the unique aspects of the image and text encoders of VLMs and further propose a self-distillation (SD) approach - VL2V-SD, to distill the superior generalization of the VLM's text encoder to its image encoder, while retaining the rich representations of the latter. We further adapt this to a black-box distillation setting, by firstly projecting the representations of the pre-trained feature-extractor (student) to a space that is aligned with embeddings of the VLM teacher, and further finetuning the backbone using the same. The use of both text and image embeddings induces the rich representations of the VLM and its superior generalization to the student model, while also making the text embeddings more suitable for use in the classification head. Both proposed approaches (VL2V-SD and VL2V-ADiP) achieve substantial gains over prior methods on the popular Domain Generalization datasets. We hope this work will motivate future work towards utilizing the uniqueness of multi-modal models effectively.

\textbf{Acknowledgements.} This work was supported by the research grant CRG/2021/005925 from SERB, DST, Govt. of India. Sravanti Addepalli is supported by the Google PhD Fellowship.

\appendix
{
    \small
    \bibliographystyle{ieeenat_fullname}
    \bibliography{main}
}
\twocolumn[
    \begin{@twocolumnfalse}
    \begin{center}
        \textbf{\huge Appendix}
        \vspace{0.5cm}
    \end{center}
    \end{@twocolumnfalse}
]
The Appendix presents further details on the proposed approach, datasets, and results. To ensure the reproducibility of our results, we share the code on our project page - \url{https://val.cds.iisc.ac.in/VL2V-ADiP/}. The Appendix is structured as follows:

\begin{itemize}
\setlength\itemsep{0.4em}
    \item \textbf{Section \ref{sec:training_algo}:} Training Algorithm
    
    \item \textbf{Section \ref{sec:datasets}:} Details on Datasets
    
    \item \textbf{Section \ref{sec:additional_results}:} Additional Results
    \begin{itemize}
        \setlength\itemsep{0.4em}
        \item \textbf{Section \ref{sec:variance}:} Variance re-runs
        \item \textbf{Section \ref{sec:swad_comparisons}:} Comparison with Additional Baselines
        \item \textbf{Section \ref{sec:other_vlms}:} Distillation using diverse VLMs
        \item \textbf{Section \ref{sec:domain-wise}:} Domain-wise Results
    \end{itemize}
    
    \item \textbf{Section \ref{sec:loss_weight}:} Analysis on Loss Weighting
\end{itemize}

\section{Training Algorithm}
\label{sec:training_algo}

The detailed training algorithm of the proposed approach VL2V-ADiP is presented in Algorithm-\ref{algo:overall}. We additionally incorporate SWAD \cite{SWAD} during training, which detects the onset of the optimal basin and performs weight-averaging across several model snapshots in the basin. To enable a fair comparison, we present results across all baselines as well using SWAD, denoted using ``(S)'' in Tables- \ref{tab:sd}, \ref{tab:main-dg}, and \ref{tab:other-students} of the main paper. 

\begin{table*}[t]
\centering
\caption{\textbf{Domain Generalization Datasets:} Details of the five DG datasets recommended by the DomainBed benchmark \cite{DomainBed}}
\vspace{-3mm}

\setlength{\extrarowheight}{1pt}%
\label{tab:datasets}
\resizebox{\textwidth}{!}{%
\begin{tabular}{@{}lcclll@{}}
\toprule
\multicolumn{1}{c}{\textbf{Dataset}} & \textbf{\begin{tabular}{c} No. of \\ classes\end{tabular}} & \textbf{\begin{tabular}{c} No. of \\ domains \end{tabular}} & \textbf{\begin{tabular}{c} No. of \\ images \end{tabular}} & \multicolumn{1}{c}{\textbf{Domains}} & \multicolumn{1}{c}{\textbf{Domain shift}} \\ \midrule

Office-Home (OH) & 65 & 4 & 15,588 & Art, Clipart, Product, Real & Style \\
Terra-Incognita (TI) & 10 & 4 & 24,788 & L100, L38, L43, L46 & Camera location \\
VLCS & 5 & 4 & 10,729 & Caltech101, LabelMe, SUN09, VOC2007 & Photography \\
PACS & 7 & 4 & 9,991 & Art, Cartoons, Photos, Sketches & Style \\
DomainNet (DN) & 345 & 6 & 586,575 & Clipart, Infograph, Painting, Quickdraw, Real, Sketch & Style \\ \bottomrule
\end{tabular}%
}
\end{table*}
\begin{table*}[t]
\centering
\caption{\textbf{Variance across re-runs:} Mean and standard deviation of the OOD accuracy (\%) of our proposed approach VL2V-ADiP when compared to the ERM and KD \cite{hinton2015distilling} baselines across the five DG datasets. All results are presented with SWAD (S) \cite{SWAD}.}
\vspace{-3mm}
\setlength{\tabcolsep}{12pt}
\setlength{\extrarowheight}{1pt}%
\label{tab:var_table}
\resizebox{\linewidth}{!}{%
\begin{tabular}{@{}lcccccc@{}}
\toprule
\multicolumn{1}{c}{\textbf{Method}} & \textbf{Office-Home} & \textbf{Terra-Incognita} & \textbf{VLCS} & \textbf{PACS} & \textbf{DomainNet} & \textbf{Avg-OOD} \\ \midrule
ERM-FFT (S) & 82.33 \std{0.87} & 48.87 \std{0.74} & 80.12 \std{0.30} & 90.15 \std{0.51} & 56.09 \std{0.09} & 71.51 \std{0.50} \\
KD (S) & 81.90 \std{0.78} & 48.90 \std{1.32} & 79.95 \std{0.49} & 90.70 \std{0.67} & 56.01 \std{0.10} & 71.49 \std{0.67} \\
VL2V-ADiP (Ours) & \textbf{85.82} \std{0.27} & \textbf{55.32} \std{0.74} & \textbf{82.31} \std{0.37} & \textbf{94.32} \std{0.56} & \textbf{59.29} \std{0.11} & \textbf{75.41} \std{0.41} \\ \bottomrule
\end{tabular}%
}
\end{table*}

\section{Details on Datasets}
\label{sec:datasets}

We evaluate the proposed approaches VL2V-SD and VL2V-ADiP on five Domain Generalization datasets that are widely used in literature and recommended on the DomainBed benchmark \cite{DomainBed}. The details of these five datasets are presented in Table-\ref{tab:datasets}. This includes diverse datasets with several unique aspects such as - less training data \cite{VLCS,PACS,OfficeHome} and a larger amount of training data \cite{DomainNet}, small domain-shifts \cite{VLCS} and larger domain shifts \cite{DomainNet}, lesser number of classes \cite{VLCS,PACS} and a higher number of classes \cite{DomainNet,OfficeHome}. We compare against several baselines on each of these individual datasets and also report the average performance across all datasets as is the standard practice \cite{DomainBed}.

\definecolor{coolblack}{rgb}{0.0, 0.23, 0.64}
\newcommand{\jnkc}[1]{\textcolor{coolblack}{#1}}
\def\code#1{\texttt{#1}}

\begin{algorithm}[t]

\caption{VL2V - Align, Distill, Predict (ADiP)}
\label{algo:overall}
\begin{algorithmic}[1]

\vspace{1mm}
\State \textbf{Input:} Let $\mathcal{D}_s=\{D_{i}, \; \forall i=1,2,\dots d-1\}$ be the data from $d-1$ source domains, $(x_i, y_i) \sim {D}_s$ be an image-label pair from source domains, $x_i^\mathrm{target}$ be a test image from the target domain, $f_{T}^{{text}}$ and $f_{T}^{{img}}$  be the text and image encoders of the VLM Teacher respectively, $f_{S}^{{fe}}$ and $f_S^{{proj}}$ be the feature extractor and linear projection layer of the student vision model respectively, $h_{\mathrm{VLM}}$ be the zero-shot classifier of the VLM teacher, and $C$ be the set of all class names in dataset $\mathcal{D}_s$. For the data sample $(x_i,y_i)$, let $\mathbf{I}_{x_i}^t$ and $\mathbf{T}_{y_i}$ be the image and text embeddings from the VLM teacher respectively, and $\textbf{PF}_{x_i}^s$ be the projected features from the student.
\vspace{1mm}

\State $P_c = \textrm{``A photo of a $c$''}$  $~ \forall~ c \in C$
\State $\mathbf{T}_c = f_T^{text}(P_c)$ $~ \forall~ c \in C$

\vspace{5mm}
\Statex \underline{{\textbf{Stage 1 - Align}}}\jnkc{\Comment{Projection layer trained}}
\vspace{2mm}
\For{$\textrm{\emph{iter}} < \textrm{\emph{MaxIters}}$}:
\State Sample batch $(x_{i}, y_{i})$ from $\mathcal{D}_s$, $\forall ~~0 \leq i < n$
\State $\textbf{I}^t_{x_i} \gets f_T^{{img}}(x_{i})$, $\forall ~~0 \leq i < n$ 
\State $\mathbf{PF}^s_{x_i} \gets f_S^{proj}(f_{S}^{fe}(x_{i}))$, $\forall ~~0 \leq i < n$ 
\State $\mathcal{L} = - \frac{1}{2n} \sum_i \big\{\cos(\textbf{PF}^s_{x_i}, \textbf{T}_{y_i}) + \cos(\textbf{PF}^s_{x_i}, \textbf{I}^t_{x_i})\big\}$
\State $\theta_{proj} \gets \theta_{proj} - \nabla_{\theta_{proj}} \mathcal{L} $
\EndFor

\vspace{5mm}
\Statex \underline{{\textbf{Stage 2 - Distill}}}\jnkc{\Comment{Feature extractor trained}}
\vspace{2mm}
\For{$\textrm{\emph{iter}} < \textrm{\emph{MaxIters}}$}:
\State Sample batch $(x_{i}, y_{i})$ from $\mathcal{D}_s$, $\forall ~~0 \leq i < n$
\State $\mathbf{I}^t_{x_i} \gets f_T^{img}(x_{i}), \forall ~~0 \leq i < n$
\State $\mathbf{PF}^S_{x_i} \gets f_S^{proj}(f_{S}^{fe}(x_{i})), \forall ~~0 \leq i < n$
\State $\mathcal{L} = - \frac{1}{2n} \sum_i \big\{\cos(\textbf{PF}^s_{x_i}, \textbf{T}_{y_i}) + \cos(\textbf{PF}^s_{x_i}, \textbf{I}^t_{x_i})\big\}$
\State $\theta_{fe} \gets \theta_{fe} - \nabla_{\theta_{fe}} \mathcal{L} $
\EndFor

\vspace{5mm}
\Statex \underline{{\textbf{Stage 3 - Predict}}}
\vspace{1mm}

\State $h_{\mathrm{VLM}}(\textbf{x}) := [~\cos(\textbf{x}, \mathbf{T}_c), ~\forall~ c \in C~]$ 
\State $\textbf{PF}^s_{x_i^\mathrm{target}} \gets f_S^{proj}(f_{S}^{fe}(x_{i}^\mathrm{target}))$
\State $\hat{y}_i = \mathrm{argmax}_c~ h_{\mathrm{VLM}}(\mathbf{PF}^s_{x_i^\mathrm{target}})$

\end{algorithmic}
\end{algorithm}

\section{Additional Results}
\label{sec:additional_results}

\subsection{Variance re-runs}
\label{sec:variance}

The results in Tables - \ref{tab:sd}, \ref{tab:main-dg}, \ref{tab:other-students}, and \ref{tab:method_abl} of the main paper are reported with a fixed seed of 0, in order to ensure reproducibility of results. In Table-\ref{tab:var_table}, we report the mean and standard deviation of the proposed method VL2V-ADiP across 3 re-runs with different random seeds. We additionally present standard deviation for the two standard baselines - ERM Fine-tuned (S) and KD (S) \cite{hinton2015distilling}, for reference. We note that the  standard deviation of the proposed method is comparable to the baselines on the respective datasets.

\begin{figure*}[t]
\centering
    \includegraphics[width=1.0\textwidth]{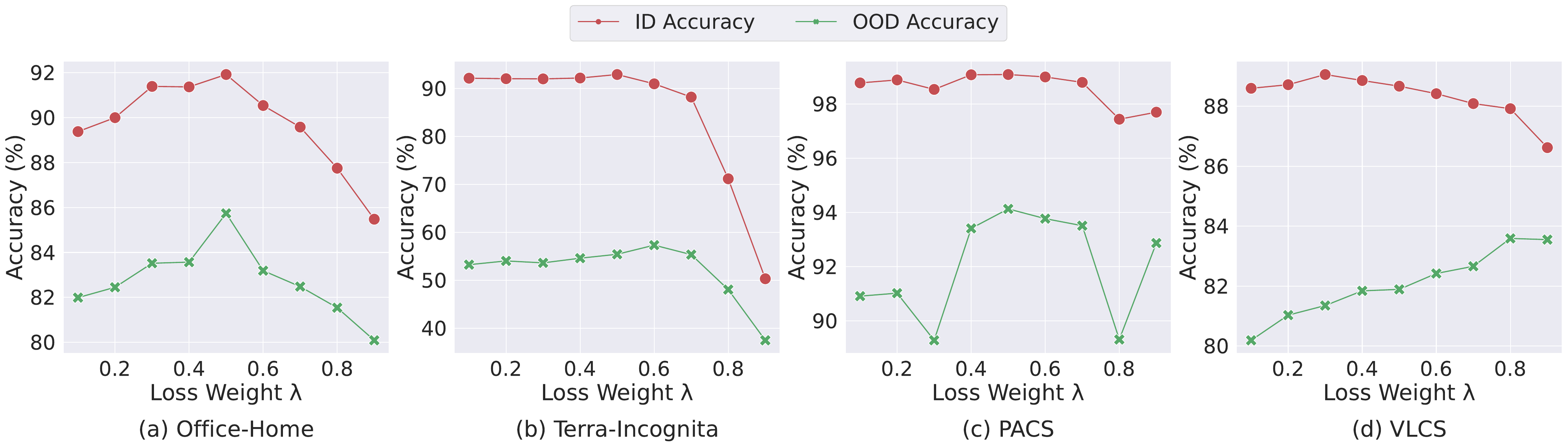}
    \caption{OOD and ID accuracy (\%) of the proposed approach VL2V-ADiP across variation in loss weight $\lambda$ for 4 Domain Generalization datasets. Cosine similarity of the student's projected features w.r.t. the text embeddings of the VLM teacher is given a weight of $(1-\lambda)$, while that w.r.t. the image embeddings of the VLM is given a weight of $\lambda$.}
    \label{fig:lambda}
    \vspace{-2mm}
\end{figure*}

\begin{table}[]
\centering
\setlength{\tabcolsep}{5pt}
\caption{\textbf{SOTA comparison with CLIP initialization (extended comparisons to show results without integrating the baselines with SWAD):} Performance (\%) of the proposed self-distillation approach VLV2-SD, compared to the SOTA DG methods. ViT-B/16 architecture is used with CLIP initialization. (S) denotes SWAD \cite{SWAD}}
\vspace{-2mm}
\label{tab:without_swad}
\resizebox{\linewidth}{!}{%
\begin{tabular}{l|ccccc|cc}
\toprule
  \textbf{Method} &
  \textbf{OH} &
  \textbf{TI} &
  \textbf{VLCS} &
  \textbf{PACS} &
  \textbf{DN} &
  \textbf{Avg-ID} &
  \textbf{Avg-OOD} \\ \midrule

  Zero-shot \cite{CLIP}   & 82.40 & 34.10          & 82.30 & 96.50 & 57.70 & -              & 70.60 \\
  SWAD \cite{SWAD}        & 81.01 & 42.92 & 79.13 & 91.35 & 57.92 & 89.05 & 70.47 \\
MIRO \cite{MIRO} & 83.36 & 54.30 & 81.32 & 95.60 & 54.00 & 89.32 & 76.32 \\
DART \cite{DART} & 77.35 & 46.41 & 77.04 & 91.45 & 56.53 & 88.65 & 69.76  \\
SAGM \cite{SAGM} & 81.11 & 54.29 & 81.11 & 90.61 & 53.59 & 89.56 & 72.41  \\
LP-FT \cite{LP-FT} & 69.72 & 36.04 & 77.10 & 86.28 & 49.00 & 84.72 & 67.14  \\
FLYP \cite{FLYP} & 75.25 & 40.22 & 75.89 & 92.97 & 48.90 & 84.66 & 69.65 \\
CLIPood \cite{CLIPood} & 67.51 & 35.68 & 78.32 & 79.61 & 47.72 & 82.52 & 65.23 \\
 RISE \cite{RISE} & 70.28 & 40.15 & 81.18 & 91.65 & 50.81 & 85.21 & 66.81 \\
\textbf{VL2V-SD (Ours)} & 85.44 & 41.18 & 82.67 & 95.67 & 58.71 & 89.50 & 72.73  \\

\midrule
\rowcolor{my_gray} \multicolumn{8}{c}{\emph{Combined with SWAD (S)} \cite{SWAD}} \\
\midrule

MIRO (S) \cite{MIRO}& 84.80 & \textbf{59.30} & 82.30 & 96.44 & 60.47 & \textbf{91.00} & 76.66 \\
DART (S) \cite{DART}& 80.93 & 51.24          & 80.38 & 93.43 & 59.32 & 89.25          & 73.06 \\
SAGM (S) \cite{SAGM}& 83.40 & 58.64          & 82.05 & 94.31 & 59.05 & 89.74          & 75.49 \\
LP-FT (S) \cite{LP-FT}& 81.17 & 47.26          & 80.88 & 92.92 & 57.04 & 88.97          & 71.85 \\
FLYP (S) \cite{FLYP}& 82.76 & 33.25          & 66.64 & 78.53 & 57.41 & 78.94          & 63.72 \\
CLIPood (S) \cite{CLIPood}& 83.31 & 46.28          & 77.19 & 93.16 & 57.78 & 69.90          & 71.55 \\
 RISE (S) \cite{RISE}& 78.39& 49.61& 80.62& 93.25& 55.37& 87.91&71.45\\
\textbf{VL2V-SD (Ours)} &
  \textbf{87.38} &
  58.54 &
  \textbf{83.25} &
  \textbf{96.68} &
  \textbf{62.79} &
  89.99 &
  \textbf{77.73} \\ \bottomrule
\end{tabular}%
}

\vspace{-4mm}
\end{table}
\begin{table}[t]
\caption{\textbf{SOTA comparison with ImageNet-1K initialization (extended comparisons to show results without integrating the baselines with SWAD):} Performance (\%) of the proposed approach VLV2-ADiP, compared to the SOTA DG methods. ViT- B/16 architecture is used with ImageNet-1K initialization. (S) denotes SWAD \cite{SWAD}}
\vspace{-2mm}

\resizebox{\linewidth}{!}{%
\label{tab:without_swad_inet}
\begin{tabular}{l|ccccc|cc}
\toprule
\textbf{Method} & \textbf{OH}    & \textbf{TI}    & \textbf{VLCS}  & \textbf{PACS}  & \textbf{DN}    & \textbf{Avg-ID} & \textbf{Avg-OOD} \\
\midrule

ERM-LP       & 71.45 & 31.80 & 77.77 & 67.52 & 36.66 & 74.25 & 57.04  \\
ERM FFT      & 78.03 & 42.53 & 78.13 & 85.32 & 50.84 & 86.90 & 66.97 \\
LP-FT \cite{LP-FT}  & 75.23 & 44.05 & 76.51 & 85.08 & 51.10 & 87.59 & 66.40 \\
SimKD \cite{SimKD} & 76.89 & 26.32 & 80.16 & 85.66 & 48.45 & 68.31 & 64.30  \\
KD    \cite{hinton2015distilling} & 77.62 & 38.66 & 79.73 & 84.87 & 50.73 & 87.04 & 66.32  \\
MIRO \cite{MIRO} & 74.88 & 44.52 & 80.39 & 81.53 & 49.95 & 86.56 & 66.25  \\
DART \cite{DART} & 82.56 & 50.70 & 79.70 & 89.76 & 56.13 & 89.99 & 71.77  \\
SAGM \cite{SAGM} & 80.87 & 52.38 & 79.53 & 87.29 & 54.04 & 88.88 & 73.83  \\
Text2Concept \cite{text2concept} & 70.57 & 26.86 & 79.03 & 66.10 & 23.29 & 53.22 & 53.17  \\
RISE \cite{RISE} & 80.34 & 44.64 & 84.15 & 90.99 & 53.29 & 87.42 & 73.47  \\
\textbf{VL2V-ADiP (Ours)} & 84.56 & 49.99 & 81.53 & 93.41 & 56.82 & 88.74 & 73.26  \\

\midrule
\rowcolor{my_gray} \multicolumn{8}{c}{\emph{Combined with SWAD (S)} \cite{SWAD}} \\
\midrule

ERM-LP (S)      & 71.48 & 31.35 & 77.52 & 67.02 & 36.65 & 73.99 & 56.81 \\
ERM FFT (S)      & 83.22 & 50.05 & 80.33 & 90.28 & 56.10 & 89.31 & 72.00 \\
LP-FT (S) \cite{LP-FT}  & 81.55	& 51.61	& 80.17	& 91.20 & 56.03 & \textbf{90.03} & 72.11 \\
SimKD (S) \cite{SimKD} & 66.76 & 81.01 & \textbf{83.92} & 28.24 & 49.42 & 68.24 & 61.87 \\
KD (S)    \cite{hinton2015distilling} & 82.73 & 48.40 & 80.48 & 91.46 & 56.11 & 89.20 & 71.84 \\
MIRO (S) \cite{MIRO} & 80.09 & 50.29 & 81.10 & 89.50 & 55.75 & 88.71 & 71.35 \\
DART (S) \cite{DART} & 83.75 & 49.68 & 77.29 & 90.55 & 58.05 & 88.54 & 71.86 \\
SAGM (S) \cite{SAGM} & 82.22 & 53.24 & 79.60 & 90.02 & 55.66 & 89.22 & 72.15 \\
Text2Concept (S) \cite{text2concept} & 70.24 & 26.46 & 64.77 & 79.03 & 23.26 & 53.15 & 52.82 \\
RISE (S) \cite{RISE} & 83.48 & 52.55 & 83.70 & 93.54 & 56.58 & 88.91 & 73.97 \\
\textbf{VL2V-ADiP (Ours)} &
  \textbf{85.74} &
  \textbf{55.43} &
  81.90&
  \textbf{94.94} &
  \textbf{59.38} &
  89.02 &
  \textbf{75.48} \\ \bottomrule
\end{tabular}%
}
\vspace{-3mm}
\end{table}
\begin{table}
\centering
\setlength{\tabcolsep}{3pt}
\caption{\textbf{Distillation to lower capacity student models:} \\Performance (\%) of the proposed approach VL2V-ADiP when compared to existing SOTA DG methods (rows), with different architectures of the student model (columns) on OfficeHome dataset. The teacher
architecture is ViT-B/16. (S) denotes SWAD.}

\label{tab:my-table}
\resizebox{\linewidth}{!}{%
\begin{tabular}{l|cccc|c}
\toprule
\textbf{Method} & \textbf{ViT-B/16} & \textbf{ViT-S/16} & \textbf{DeiT-S/16} & \textbf{ResNet-50} & \textbf{Avg.} \\
\midrule

ERM-LP (S) & 71.48& 68.47& 74.12& 68.46& 70.63\\
ERM-FFT (S)&  83.22&  78.58&  74.95&  70.85&  76.90\\
 LP-FT (S) \cite{LP-FT}& 81.55& 78.77& 74.41& 70.39&76.28\\
SimKD (S) \cite{SimKD} & 66.76 & 54.18 & 58.75 & 60.88 & 60.14 \\
KD (S) \cite{hinton2015distilling} & 82.73 & 78.14 & 74.65 & 70.67 & 76.55 \\
MIRO (S) \cite{MIRO} & 80.09 & 69.45 & 73.18 & 72.40 & 73.78 \\
DART (S) \cite{DART} & 83.75 & 79.67 & 75.85 & 71.90 & 77.79 \\
SAGM (S) \cite{SAGM} & 82.22 & 77.00 & 73.94 & 70.10 & 75.81 \\
Text2Concept (S) \cite{text2concept} &  70.24&  63.30&  66.27&  61.89&  65.42\\
RISE (S) \cite{RISE} &  83.48&  80.47&  76.09&  72.40&  78.11\\
\textbf{VL2V-ADiP (Ours)} & \textbf{85.74} & \textbf{81.22} & \textbf{77.63} & \textbf{74.42} & \textbf{79.75} \\

\bottomrule

\end{tabular}%
}
\end{table}

\subsection{Comparison with Additional Baselines}
\label{sec:swad_comparisons}

We present additional baseline results corresponding to Tables - \ref{tab:sd}, \ref{tab:main-dg}, and \ref{tab:other-students} of the main paper in Tables-\ref{tab:without_swad}, \ref{tab:without_swad_inet} and \ref{tab:my-table} respectively, for the sake of completeness. In Tables-\ref{tab:without_swad} and \ref{tab:without_swad_inet}, we additionally present the respective baseline results without including SWAD \cite{SWAD} during training. In Table-\ref{tab:my-table}, we compare the performance of the proposed approach VL2V-ADiP on the OfficeHome dataset, with all the baselines considered in Table-\ref{tab:main-dg} of the main paper, on student models with different architectures. The proposed approaches show gains across baselines in all the tables.

\subsection{Distillation using diverse VLMs}
\label{sec:other_vlms}
We demonstrate the compatibility of the proposed method VL2V-ADiP with diverse VLM teacher models in Table-\ref{tab:other_vlms}. Specifically, we show results by distilling from FLAVA \cite{FLAVA}, BLIP \cite{BLIP}, and the data-efficient versions \cite{DeCLIP} of CLIP and FILIP \cite{yao2021filip}. We observe that our method achieves the highest gains over the KD baseline \cite{hinton2015distilling} with CLIP, where the teacher VLM has been trained with a large pre-training dataset. However, our method achieves significant gains even with VLMs pre-trained on smaller datasets.

\begin{table}[t]
\centering
\caption{\textbf{Distillation using various VLMs:} Performance (\%) of the proposed approach VL2V-ADiP (denoted as Ours) on 4 DG datasets, when distilling from FLAVA \cite{FLAVA}, BLIP \cite{BLIP}, CLIP \cite{CLIP} and the data-efficient versions \cite{DeCLIP} of CLIP and FILIP \cite{yao2021filip}. The student architecture is ViT-B/16 in all cases.}
\label{tab:other_vlms}
\resizebox{\linewidth}{!}{%
\begin{tabular}{@{}ccl|cccc|c@{}}
\toprule
\textbf{Teacher} &
\textbf{Dataset} &
\textbf{Method} &
\textbf{OH}&
\textbf{VLCS} &
\textbf{PACS} &
\textbf{TI}&
\textbf{Avg-OOD}
\\ \midrule

\multirow{3}{*}{\begin{tabular}[c]{@{}c@{}}FLAVA\\ ViT-B/16\end{tabular}} &
\multirow{3}{*}{\begin{tabular}[c]{@{}c@{}}PMD \\ 70M\end{tabular}}&
  Zero-shot &
  69.99 &
  79.21 &
  91.34 &
  28.85 &
  67.35 \\
 & &
  KD (S) &
  82.50 &
  80.41 &
  90.71 &
  50.86 &
  76.12 \\
 & &
  \textbf{Ours} &
  \textbf{84.16} &
  \textbf{82.94} &
  \textbf{93.22} &
  \textbf{54.56} &
  \textbf{78.72} \\ 
  \midrule
\multirow{3}{*}{\begin{tabular}[c]{@{}c@{}}BLIP\\ ViT-B/16\end{tabular}} &
\multirow{3}{*}{\begin{tabular}[c]{@{}c@{}}CapFilt \\ 129M\end{tabular}} &
  Zero-shot &
  84.83 &
  71.60 &
  92.23 &
  29.75 &
  69.60 \\
 & &
  KD (S) &
  82.45 &
  80.31 &
  87.73 &
  48.03 &
  74.63 \\
 & &
  \textbf{Ours} &
  \textbf{85.86} &
  \textbf{81.60} &
  \textbf{94.10} &
  \textbf{52.07} &
  \textbf{78.41} \\
  \midrule
\multirow{3}{*}{\begin{tabular}[c]{@{}c@{}}CLIP\\ ViT-B/16\end{tabular}} &
\multirow{3}{*}{\begin{tabular}[c]{@{}c@{}}CLIP \\ 400M\end{tabular}} &
  Zero-shot &
  81.57 &
  82.55 &
  95.99 &
  31.15 &
  72.81 \\
 & &
  KD (S) &
  82.73 &
  80.48 &
  91.49 &
  48.33 &
  75.76 \\
 & &
  \textbf{Ours} &
  \textbf{85.74} &
  \textbf{81.89} &
  \textbf{94.13} &
  \textbf{55.43} &
  \textbf{79.30} \\ \midrule
\multirow{3}{*}{\begin{tabular}[c]{@{}c@{}}DeCLIP\\ ViT-B/32\end{tabular}} &
\multirow{3}{*}{\begin{tabular}[c]{@{}c@{}}YFCC \\ 15M\end{tabular}} &
  Zero-shot &
  43.46 &
  77.79 &
  83.69 &
  27.70 &
  58.16 \\
 & &
  KD (S) &
  81.84 &
  79.95 &
  89.96 &
  49.49 &
  75.31 \\
 & &
  \textbf{Ours} &
  \textbf{82.85} &
  \textbf{81.40} &
  \textbf{92.16} &
  \textbf{50.50} &
  \textbf{76.73} \\
  \midrule
\multirow{3}{*}{\begin{tabular}[c]{@{}c@{}}DeFILIP\\ ViT-B/32\end{tabular}} &
\multirow{3}{*}{\begin{tabular}[c]{@{}c@{}}YFCC \\ 15M\end{tabular}} &
  Zero-shot &
  46.97 &
  74.08 &
  82.02 &
  16.34 &
  54.85 \\
 & &
  KD (S) &
  82.14 &
  79.53 &
  90.68 &
  50.96 &
  75.83 \\
 & &
  \textbf{Ours} &
  \textbf{83.11} &
  \textbf{81.43} &
  \textbf{92.03} &
  \textbf{51.69} &
  \textbf{77.06} \\
  \bottomrule
\end{tabular}%
}
\vspace{-3mm}
\end{table}

\subsection{Domain-wise Results}
\label{sec:domain-wise}

\begin{table*}[t]
\centering
\caption{\textbf{Domain-wise performance} (\%) of the proposed approaches VL2V-SD and VL2V-ADiP when compared to the respective baselines, on individual domains of all Domain Generalization datasets on the DomainBed benchmark \cite{DomainBed}.}
\vspace{-4mm}
\label{tab:dom_wise_res}
\resizebox{\textwidth}{!}{%
\begin{tabular}{cc}

\begin{subtable}{0.48\textwidth}
\centering
\caption{Domain-wise OOD accuracy for the approach VL2V-SD compared to the baselines combined with SWAD (S) \cite{SWAD} for the white box setting.}
\label{tab:sd_dom_split}
\resizebox{\columnwidth}{!}{%
\begin{tabular}{lrcccclc}
\toprule
\multicolumn{1}{c}{\textbf{Method / Dataset}} &
  \multicolumn{7}{c}{\textbf{Domains}} \\\midrule
\multicolumn{1}{c}{\textit{OfficeHome}} &
  \multicolumn{2}{r}{\textit{Art}} &
  \textit{Clipart} &
  \textit{Product} &
  \textit{Real} &
  \multicolumn{2}{l}{\textit{Avg.}} \\\midrule
ERM Full Fine-Tuning (S) &
  \multicolumn{2}{r}{80.12} &
  70.25 &
  86.18 &
  87.49 &
  \multicolumn{2}{l}{81.01} \\
MIRO (S) \cite{MIRO} &
  \multicolumn{2}{r}{83.57} &
  75.72 &
  89.70 &
  90.22 &
  \multicolumn{2}{l}{84.80} \\
DART (S) \cite{DART} &
  \multicolumn{2}{r}{78.79} &
  72.71 &
  86.04 &
  86.17 &
  \multicolumn{2}{l}{80.93} \\
SAGM (S) \cite{SAGM} &
  \multicolumn{2}{r}{82.60} &
  72.94 &
  88.94 &
  89.13 &
  \multicolumn{2}{l}{83.40} \\
LP-FT (S) \cite{LP-FT} &
  \multicolumn{2}{r}{80.18} &
  71.94 &
  86.35 &
  86.23 &
  \multicolumn{2}{l}{81.17} \\
CLIPood (S) \cite{CLIPood} &
  \multicolumn{2}{r}{84.86} &
  70.93 &
  88.09 &
  89.39 &
  \multicolumn{2}{l}{83.31} \\
RISE (S) \cite{RISE} &
  \multicolumn{2}{r}{75.08} &
  69.16 &
  84.35 &
  84.97 &
  \multicolumn{2}{l}{78.39} \\
WiSE-FT \cite{WISE-FT} &
  \multicolumn{2}{r}{85.15} &
  76.17 &
  \textbf{92.90} &
  91.04 &
  \multicolumn{2}{l}{86.32} \\
\textbf{VL2V-SD (Ours)} &
  \multicolumn{2}{r}{\textbf{87.33}} &
  \textbf{78.55} &
  91.98 &
  \textbf{91.65} &
  \multicolumn{2}{l}{\textbf{87.38}} \\\midrule
\multicolumn{1}{c}{\textit{TerraIncognita}} &
  \multicolumn{2}{r}{\textit{L100}} &
  \textit{L38} &
  \textit{L43} &
  \textit{L46} &
  \multicolumn{2}{l}{\textit{Avg.}} \\\midrule
ERM Full Fine-Tuning (S) &
  \multicolumn{2}{r}{38.94} &
  38.03 &
  54.03 &
  40.68 &
  \multicolumn{2}{l}{42.92} \\
MIRO (S) \cite{MIRO} &
  \multicolumn{2}{r}{67.15} &
  50.75 &
  \textbf{66.63} &
  52.71 &
  \multicolumn{2}{l}{\textbf{59.30}} \\
DART (S) \cite{DART} &
  \multicolumn{2}{r}{61.09} &
  39.48 &
  58.97 &
  45.42 &
  \multicolumn{2}{l}{51.24} \\
SAGM (S) \cite{SAGM} &
  \multicolumn{2}{r}{\textbf{72.21}} &
  50.10 &
  62.50 &
  49.73 &
  \multicolumn{2}{l}{58.64} \\
LP-FT (S) \cite{LP-FT} &
  \multicolumn{2}{r}{54.23} &
  38.09 &
  56.52 &
  40.20 &
  \multicolumn{2}{l}{47.26} \\
CLIPood (S) \cite{CLIPood} &
  \multicolumn{2}{r}{47.32} &
  38.12 &
  55.73 &
  43.94 &
  \multicolumn{2}{l}{46.28} \\
RISE (S) \cite{RISE} &
  \multicolumn{2}{r}{60.30} &
  37.43 &
  56.77 &
  43.94 &
  \multicolumn{2}{l}{49.61} \\
WiSE-FT \cite{WISE-FT} &
  \multicolumn{2}{r}{56.75} &
  \textbf{51.93} &
  61.71 &
  47.62 &
  \multicolumn{2}{l}{54.50} \\
\textbf{VL2V-SD (Ours)} &
  \multicolumn{2}{r}{69.10} &
  48.40 &
  63.10 &
  \textbf{53.56} &
  \multicolumn{2}{l}{58.54} \\\midrule
\multicolumn{1}{c}{\textit{VLCS}} &
  \multicolumn{2}{r}{\textit{Caltech}} &
  \textit{LabelMe} &
  \textit{Pascal} &
  \textit{Sun} &
  \multicolumn{2}{l}{\textit{Avg.}} \\\midrule
ERM Full Fine-Tuning (S) &
  \multicolumn{2}{r}{99.12} &
  63.31 &
  79.01 &
  75.08 &
  \multicolumn{2}{l}{79.13} \\
MIRO (S) \cite{MIRO} &
  \multicolumn{2}{r}{97.53} &
  66.59 &
  81.57 &
  \textbf{83.53} &
  \multicolumn{2}{l}{82.30} \\
DART (S) \cite{DART} &
  \multicolumn{2}{r}{99.12} &
  65.73 &
  81.07 &
  75.63 &
  \multicolumn{2}{l}{80.38} \\
SAGM (S) \cite{SAGM} &
  \multicolumn{2}{r}{97.73} &
  65.86 &
  83.77 &
  80.85 &
  \multicolumn{2}{l}{82.05} \\
LP-FT (S) \cite{LP-FT} &
  \multicolumn{2}{r}{98.24} &
  65.66 &
  81.23 &
  78.40 &
  \multicolumn{2}{l}{80.88} \\
CLIPood (S) \cite{CLIPood} &
  \multicolumn{2}{r}{82.22} &
  66.47 &
  84.19 &
  75.90 &
  \multicolumn{2}{l}{77.19} \\
RISE (S) \cite{RISE} &
  \multicolumn{2}{r}{\textbf{99.50}} &
  67.27 &
  81.44 &
  74.27 &
  \multicolumn{2}{l}{80.62} \\
WiSE-FT \cite{WISE-FT} &
  \multicolumn{2}{r}{98.99} &
  66.04 &
  83.47 &
  83.01 &
  \multicolumn{2}{l}{82.88} \\
\textbf{VL2V-SD (Ours)} &
  \multicolumn{2}{r}{99.24} &
  \textbf{67.81}&
  \textbf{86.89} &
  79.05 &
  \multicolumn{2}{l}{\textbf{83.25}} \\\midrule
\multicolumn{1}{c}{\textit{PACS}} &
  \multicolumn{2}{r}{\textit{Art}} &
  \textit{Cartoon} &
  \textit{Photo} &
  \textit{Sketch} &
  \multicolumn{2}{l}{\textit{Avg.}} \\\midrule
ERM Full Fine-Tuning (S) &
  \multicolumn{2}{r}{91.34} &
  89.07 &
  97.53 &
  87.47 &
  \multicolumn{2}{l}{91.35} \\
MIRO (S) \cite{MIRO} &
  \multicolumn{2}{r}{98.05} &
  97.50 &
  99.78 &
  90.46 &
  \multicolumn{2}{l}{96.44} \\
DART (S) \cite{DART} &
  \multicolumn{2}{r}{94.45} &
  92.27 &
  98.80 &
  88.20 &
  \multicolumn{2}{l}{93.43} \\
SAGM (S) \cite{SAGM} &
  \multicolumn{2}{r}{95.18} &
  93.60 &
  99.03 &
  89.41 &
  \multicolumn{2}{l}{94.31} \\
LP-FT (S) \cite{LP-FT} &
  \multicolumn{2}{r}{91.46} &
  92.59 &
  99.10 &
  88.52 &
  \multicolumn{2}{l}{92.92} \\
CLIPood (S) \cite{CLIPood} &
  \multicolumn{2}{r}{92.68} &
  91.05 &
  98.95 &
  89.98 &
  \multicolumn{2}{l}{93.16} \\
RISE (S) \cite{RISE} &
  \multicolumn{2}{r}{92.25} &
  93.82 &
  98.65 &
  88.26 &
  \multicolumn{2}{l}{93.25} \\
WiSE-FT \cite{WISE-FT} &
  \multicolumn{2}{r}{\textbf{98.29}} &
  \textbf{98.50}&
  \textbf{100.00} &
  \textbf{92.36} &
  \multicolumn{2}{l}{\textbf{97.29}} \\
\textbf{VL2V-SD (Ours)} &
  \multicolumn{2}{r}{98.05} &
  98.19 &
  99.93 &
  90.55 &
  \multicolumn{2}{l}{96.68} \\\midrule
\multicolumn{1}{c}{\textit{DomainNet}} &
  \multicolumn{1}{c}{\textit{clp}} &
  \textit{inf} &
  \textit{pnt} &
  \textit{qkdr} &
  \textit{real} &
  \multicolumn{1}{c}{\textit{skt}} &
  \textit{Avg.} \\\midrule
ERM Full Fine-Tuning (S) &
  \multicolumn{1}{c}{77.10} &
  38.32 &
  66.13 &
  25.02 &
  75.19 &
  \multicolumn{1}{c}{65.76} &
  57.92 \\
MIRO (S) \cite{MIRO} &
  \multicolumn{1}{c}{79.70} &
  43.50 &
  67.36 &
  24.62 &
  79.22 &
  \multicolumn{1}{c}{68.42} &
  60.47 \\
DART (S) \cite{DART} &
  \multicolumn{1}{c}{78.51} &
  39.99 &
  66.89 &
  \textbf{25.85}&
  76.37 &
  \multicolumn{1}{c}{68.29} &
  59.32 \\
SAGM (S) \cite{SAGM} &
  \multicolumn{1}{c}{78.78} &
  40.21 &
  67.31 &
  24.18 &
  76.29 &
  \multicolumn{1}{c}{67.54} &
  59.05 \\
LP-FT (S) \cite{LP-FT} &
  \multicolumn{1}{c}{77.37} &
  33.88 &
  65.27 &
  24.82 &
  74.60 &
  \multicolumn{1}{c}{66.32} &
  57.04 \\
CLIPood (S) \cite{CLIPood} &
  \multicolumn{1}{c}{76.28} &
  38.46 &
  66.98 &
  21.76 &
  75.79 &
  \multicolumn{1}{c}{67.43} &
  57.78 \\
RISE (S) \cite{RISE} &
  \multicolumn{1}{c}{77.80} &
  31.32 &
  57.64 &
  24.60 &
  73.96 &
  \multicolumn{1}{c}{66.90} &
  55.37 \\
WiSE-FT \cite{WISE-FT} &
  \multicolumn{1}{c}{72.74} &
  46.36 &
  64.05 &
  16.79 &
  \textbf{82.82}&
  \multicolumn{1}{c}{65.29} &
  58.01 \\
\textbf{VL2V-SD (Ours)} &
  \multicolumn{1}{c}{\textbf{79.96}} &
  \textbf{49.00} &
  \textbf{71.05} &
  23.34 &
  82.05 &
  \multicolumn{1}{c}{\textbf{71.36}} &
  \textbf{62.79} \\
\bottomrule
\end{tabular}%
}
\label{tab:oh_dom_split_sd}
\end{subtable} &

\begin{subtable}{0.48\textwidth}
\centering
\caption{Domain-wise OOD accuracy for the approach VL2V-ADiP compared to the baselines combined with SWAD (S) \cite{SWAD} for the black box setting.}
\label{tab:ad_dom_split}
\resizebox{\columnwidth}{!}{%
\begin{tabular}{lrcccclc}
\toprule
\multicolumn{1}{c}{\textbf{Method / Dataset}} &
  \multicolumn{7}{c}{\textbf{Domains}} \\ \midrule
\multicolumn{1}{c}{\textit{OfficeHome}} &
  \multicolumn{2}{r}{\textit{Art}} &
  \textit{Clipart} &
  \textit{Product} &
  \textit{Real} &
  \multicolumn{2}{l}{\textit{Avg.}} \\ \midrule
ERM Full Fine-Tuning (S) &
  \multicolumn{2}{r}{82.24} &
  72.19 &
  88.43 &
  90.02 &
  \multicolumn{2}{l}{83.22} \\
LP-FT (S) \cite{LP-FT} &
  \multicolumn{2}{r}{81.67} &
  65.64 &
  89.22 &
  89.67 &
  \multicolumn{2}{l}{81.55} \\
KD (S) \cite{hinton2015distilling} &
  \multicolumn{2}{r}{80.79} &
  70.76 &
  89.08 &
  90.30 &
  \multicolumn{2}{l}{82.73} \\
MIRO (S) \cite{MIRO} &
  \multicolumn{2}{r}{78.89} &
  64.15 &
  87.70 &
  89.62 &
  \multicolumn{2}{l}{80.09} \\
DART (S) \cite{DART} &
  \multicolumn{2}{r}{81.72} &
  73.17 &
  89.64 &
  90.48 &
  \multicolumn{2}{l}{83.75} \\
SAGM (S) \cite{SAGM} &
  \multicolumn{2}{r}{80.23} &
  70.16 &
  88.46 &
  90.02 &
  \multicolumn{2}{l}{82.22} \\
Text2Concept (S) \cite{text2concept} &
  \multicolumn{2}{r}{71.06} &
  48.97 &
  77.48 &
  83.45 &
  \multicolumn{2}{l}{70.24} \\
RISE (S) \cite{RISE} &
  \multicolumn{2}{r}{81.87} &
  72.42 &
  89.27 &
  90.33 &
  \multicolumn{2}{l}{83.48} \\
\textbf{VL2V-ADiP (Ours)} &
  \multicolumn{2}{r}{\textbf{84.81}} &
  \textbf{75.92} &
  \textbf{90.65} &
  \textbf{91.60} &
  \multicolumn{2}{l}{\textbf{85.74}} \\ \midrule
\multicolumn{1}{c}{\textit{TerraIncognita}} &
  \multicolumn{2}{r}{\textit{L100}} &
  \textit{L38} &
  \textit{L43} &
  \textit{L46} &
  \multicolumn{2}{l}{\textit{Avg.}} \\ \midrule
ERM Full Fine-Tuning (S) &
  \multicolumn{2}{r}{58.98} &
  37.76 &
  58.31 &
  45.15 &
  \multicolumn{2}{l}{50.05} \\
LP-FT (S) \cite{LP-FT} &
  \multicolumn{2}{r}{58.29} &
  41.08 &
  \textbf{63.22} &
  43.83 &
  \multicolumn{2}{l}{51.61} \\
KD (S) \cite{hinton2015distilling} &
  \multicolumn{2}{r}{61.09} &
  33.21 &
  57.84 &
  41.17 &
  \multicolumn{2}{l}{48.33} \\
MIRO (S) \cite{MIRO} &
  \multicolumn{2}{r}{61.27} &
  38.14 &
  57.68 &
  44.08 &
  \multicolumn{2}{l}{50.29} \\
DART (S) \cite{DART} &
  \multicolumn{2}{r}{56.82} &
  37.34 &
  62.31 &
  42.26 &
  \multicolumn{2}{l}{49.68} \\
SAGM (S) \cite{SAGM} &
  \multicolumn{2}{r}{\textbf{64.17}} &
  44.42 &
  59.64 &
  44.74 &
  \multicolumn{2}{l}{53.24} \\
Text2Concept (S) \cite{text2concept} &
  \multicolumn{2}{r}{43.55} &
  2.04 &
  31.83 &
  28.43 &
  \multicolumn{2}{l}{26.46} \\
RISE (S) \cite{RISE} &
  \multicolumn{2}{r}{59.77} &
  43.79 &
  59.45 &
  47.19 &
  \multicolumn{2}{l}{52.55} \\
\textbf{VL2V-ADiP (Ours)} &
  \multicolumn{2}{r}{62.93} &
  \textbf{44.83} &
  60.71 &
  \textbf{53.26} &
  \multicolumn{2}{l}{\textbf{55.43}} \\ \midrule
\multicolumn{1}{c}{\textit{VLCS}} &
  \multicolumn{2}{r}{\textit{Caltech}} &
  \textit{LabelMe} &
  \textit{Pascal} &
  \textit{Sun} &
  \multicolumn{2}{l}{\textit{Avg.}} \\ \midrule
ERM Full Fine-Tuning (S) &
  \multicolumn{2}{r}{98.49} &
  64.05 &
  82.60 &
  76.17 &
  \multicolumn{2}{l}{80.33} \\
LP-FT (S) \cite{LP-FT} &
  \multicolumn{2}{r}{96.97} &
  63.58 &
  82.02 &
  78.13 &
  \multicolumn{2}{l}{80.17} \\
KD (S) \cite{hinton2015distilling} &
  \multicolumn{2}{r}{98.87} &
  65.32 &
  81.33 &
  76.39 &
  \multicolumn{2}{l}{80.48} \\
MIRO (S) \cite{MIRO} &
  \multicolumn{2}{r}{99.75} &
  64.79 &
  82.66 &
  77.20 &
  \multicolumn{2}{l}{81.10} \\
DART (S) \cite{DART} &
  \multicolumn{2}{r}{94.08} &
  63.11 &
  76.12 &
  75.86 &
  \multicolumn{2}{l}{77.29} \\
SAGM (S) \cite{SAGM} &
  \multicolumn{2}{r}{98.49} &
  64.92 &
  79.32 &
  75.68 &
  \multicolumn{2}{l}{79.60} \\
Text2Concept (S) \cite{SAGM} &
  \multicolumn{2}{r}{98.36} &
  68.08 &
  77.21 &
  72.47 &
  \multicolumn{2}{l}{79.03} \\
RISE (S) \cite{RISE} &
  \multicolumn{2}{r}{\textbf{100.00}} &
  \textbf{69.15}&
  \textbf{84.03}&
  \textbf{81.61} &
  \multicolumn{2}{l}{\textbf{83.70}} \\
\textbf{VL2V-ADiP (Ours)} &
  \multicolumn{2}{r}{99.62} &
  66.60 &
  82.87 &
  78.46 &
  \multicolumn{2}{l}{81.89} \\ \midrule
\multicolumn{1}{c}{\textit{PACS}} &
  \multicolumn{2}{r}{\textit{Art}} &
  \textit{Cartoon} &
  \textit{Photo} &
  \textit{Sketch} &
  \multicolumn{2}{l}{\textit{Avg.}} \\ \midrule
ERM Full Fine-Tuning (S) &
  \multicolumn{2}{r}{93.78} &
  86.25 &
  99.18 &
  81.93 &
  \multicolumn{2}{l}{90.28} \\
LP-FT (S) \cite{LP-FT} &
  \multicolumn{2}{r}{94.27} &
  86.83 &
  99.48 &
  84.22 &
  \multicolumn{2}{l}{91.20} \\
KD (S) \cite{hinton2015distilling} &
  \multicolumn{2}{r}{94.20} &
  86.35 &
  99.25 &
  86.04 &
  \multicolumn{2}{l}{91.46} \\
MIRO (S) \cite{MIRO} &
  \multicolumn{2}{r}{94.69} &
  85.98 &
  99.63 &
  77.70 &
  \multicolumn{2}{l}{89.50} \\
DART (S) \cite{DART} &
  \multicolumn{2}{r}{94.45} &
  86.67 &
  99.55 &
  81.52 &
  \multicolumn{2}{l}{90.55} \\
SAGM (S) \cite{SAGM} &
  \multicolumn{2}{r}{93.72} &
  86.57 &
  99.18 &
  80.63 &
  \multicolumn{2}{l}{90.02} \\
Text2Concept (S) \cite{text2concept} &
  \multicolumn{2}{r}{80.17} &
  66.47 &
  96.63 &
  15.81 &
  \multicolumn{2}{l}{64.77} \\
RISE (S) \cite{RISE} &
  \multicolumn{2}{r}{93.72} &
  \textbf{93.23}&
  99.55 &
  87.66 &
  \multicolumn{2}{l}{93.54} \\
\textbf{VL2V-ADiP (Ours)} &
  \multicolumn{2}{r}{\textbf{95.61}} &
  92.38 &
  \textbf{99.85}&
  \textbf{88.68}&
  \multicolumn{2}{l}{\textbf{94.13}} \\ \midrule
\multicolumn{1}{c}{\textit{DomainNet}} &
  \multicolumn{1}{c}{\textit{clp}} &
  \textit{inf} &
  \textit{pnt} &
  \textit{qkdr} &
  \textit{real} &
  \multicolumn{1}{c}{\textit{skt}} &
  \textit{Avg.} \\ \midrule
ERM Full Fine-Tuning (S) &
  \multicolumn{1}{c}{76.34} &
  30.92 &
  64.76 &
  21.30 &
  77.70 &
  \multicolumn{1}{c}{65.60} &
  56.10 \\
LP-FT (S) \cite{LP-FT} &
  \multicolumn{1}{c}{76.49} &
  30.75 &
  65.30 &
  20.82 &
  77.83 &
  \multicolumn{1}{c}{64.98} &
  56.03 \\
KD (S) \cite{hinton2015distilling} &
  \multicolumn{1}{c}{76.56} &
  31.29 &
  64.55 &
  \textbf{21.44}&
  77.62 &
  \multicolumn{1}{c}{65.23} &
  56.11 \\
MIRO (S) \cite{MIRO} &
  \multicolumn{1}{c}{76.32} &
  30.96 &
  64.52 &
  20.18 &
  77.88 &
  \multicolumn{1}{c}{64.63} &
  55.75 \\
DART (S) \cite{DART} &
  \multicolumn{1}{c}{77.62} &
  34.14 &
  67.64 &
  21.05 &
  80.73 &
  \multicolumn{1}{c}{67.11} &
  58.05 \\
SAGM (S) \cite{SAGM} &
  \multicolumn{1}{c}{76.67} &
  29.85 &
  64.42 &
  20.68 &
  77.58 &
  \multicolumn{1}{c}{64.58} &
  55.63 \\
Text2Concept (S) \cite{text2concept} &
  \multicolumn{1}{c}{22.41} &
  10.14 &
  35.26 &
  0.47 &
  56.55 &
  \multicolumn{1}{c}{14.61} &
  23.26 \\
RISE (S) \cite{RISE} &
  \multicolumn{1}{c}{77.87} &
  32.71 &
  61.03 &
  21.20 &
  79.90 &
  \multicolumn{1}{c}{66.77} &
  56.58 \\
\textbf{VL2V-ADiP (Ours)} &
  \multicolumn{1}{c}{\textbf{78.80}} &
  \textbf{36.86}&
  \textbf{69.21}&
  21.32 &
  \textbf{81.33}&
  \multicolumn{1}{c}{\textbf{68.79}} &
  \textbf{59.38}\\ \bottomrule
\end{tabular}%
}
\label{tab:oh_dom_split_ad}
\end{subtable} \\

\end{tabular}%
}
\end{table*}

We present the results of the proposed approaches VL2V-SD and VL2V-ADiP on each of the individual domains in Table-\ref{tab:sd_dom_split} and Table-\ref{tab:ad_dom_split}. The domain in the column heading indicates the unseen test domain, where the training was done on the remaining $d-1$ domains mentioned in Table-\ref{tab:datasets}. We note that the proposed methods VL2V-SD and VL2V-ADiP outperform existing methods across several datasets and domains. 

VL2V-ADiP achieves the highest gains in cases where domain shift is large, highlighting the benefit of using the supervision from CLIP in improving OOD generalization on downstream tasks. The domains with the highest gains include ClipArt (OH), Location-38 (TI), Location-46 (TI), Cartoon (PACS), Infograph (DN), and Painting (OH). The domains with the least gains include Product (OH), Real-World (OH), Art (PACS), Photo (PACS), Quickdraw (DN), and all domains in VLCS. It is intuitive to see that most of the domains with the least gains are the cases where the target distribution is similar to at least one of the source distributions, making them less challenging to evaluate OOD robustness. For example, there is no real domain shift in VLCS, apart from the fact that each split is obtained from a different dataset, with a possible domain shift due to photography differences, which can be considered minor. Hence, taking the supervision of a CLIP model is the least beneficial here. 

\section{Analysis on Loss Weighting}
\label{sec:loss_weight}

The training loss of the proposed approach VL2V-ADiP presented in Eq.\ref{eq:vl2v-adip} of the main paper, and in L8 and L15 of Algorithm-\ref{algo:overall}, considers equal weights on both loss terms - cosine similarity of the image embeddings $\mathbf{PF}^s_{x_i}$ w.r.t. text and image embeddings of the VLM teacher respectively. In this section, we explore the impact of varying these weights as a convex interpolation between the cosine similarity w.r.t. text embeddings (weighted by $1-\lambda$) and image embeddings (weighted by $\lambda$) respectively as shown below:

\begin{equation}
\label{eq:vl2v-adip-lambda}
    \mathcal{L} = - \frac{1}{2n} \sum_{i=1}^n \big\{(1-\lambda) \cdot \cos(\textbf{PF}^s_{x_i}, \textbf{T}_{y_i}) + \lambda \cdot \cos(\textbf{PF}^s_{x_i}, \textbf{I}^t_{x_i})\big\}
\end{equation}

We note from the plots in Fig.\ref{fig:lambda} that while the best OOD accuracy could be achieved at a different $\lambda$ value, a setting of 0.5 works reasonably well, since the proposed approach is not too sensitive to variations in $\lambda$ in most cases. Moreover, a value of 0.5 assigns equal weightage to losses w.r.t. both image and text embeddings (since they are of the same scale), which is the best setting to consider in the absence of hyperparameter tuning.

\end{document}